\pdfoutput=1

\documentclass[11pt]{article}

\usepackage[final]{acl}

\usepackage{times}
\usepackage{latexsym}

\usepackage[T1]{fontenc}

\usepackage[utf8]{inputenc}

\usepackage{microtype}

\usepackage{inconsolata}

\usepackage{graphicx}
\usepackage{amsmath}
\usepackage{dsfont}
\usepackage{adjustbox}
\usepackage{subcaption}
\usepackage{amssymb}
\usepackage{booktabs}
\usepackage{multirow}
\usepackage{pifont}
\usepackage[table]{xcolor}
\definecolor{lightblue}{rgb}{0.84,0.91,0.97}
\definecolor{red}{rgb}{0.86, 0.08, 0.24}

\title{Captioning for Text-Video Retrieval via \\ Dual-Group Direct Preference Optimization}

\author{
Ji Soo Lee\textsuperscript{\rm 1}\hspace{0.3cm}
Byungoh Ko\textsuperscript{\rm 1}\hspace{0.3cm} 
Jaewon Cho\textsuperscript{\rm 1}\hspace{0.3cm}
Howoong Lee\textsuperscript{\rm 2}\hspace{0.3cm}
Jaewoon Byun\textsuperscript{\rm 2}\hspace{0.3cm}
Hyunwoo J. Kim\textsuperscript{\rm 3}$^{\dagger}$\vspace{0.3cm} \\
\textsuperscript{\rm 1}Korea University\hspace{0.8cm}
\textsuperscript{\rm 2}Hanwha Vision\hspace{0.8cm}
\textsuperscript{\rm 3}KAIST\vspace{0.3cm} \\
\tt\small \{simplewhite9, ko990128, cho35750\}@korea.ac.kr \vspace{0cm} \\
\tt\small \{howoong.lee, jaewoon.byun\}@hanwha.com \hspace{0.5cm} 
\tt\small hyunwoojkim@kaist.ac.kr
}

\begin{document}
\maketitle
\renewcommand{\thefootnote}{\fnsymbol{footnote}}
\footnotetext[0]{$\dagger$ Corresponding authors.} 
\begin{abstract}
In text-video retrieval, auxiliary captions are often used to enhance video understanding, bridging the gap between the modalities.
While recent advances in multi-modal large language models (MLLMs) have enabled strong zero-shot caption generation, we observe that such captions tend to be generic and indistinguishable across visually similar videos, limiting their utility for fine-grained retrieval. 
Moreover, conventional captioning approaches are typically evaluated using language generation metrics, such as BLEU, which are not typically tailored for retrieval tasks that require making discriminative distinctions between candidates.
To address this, we propose \textbf{CaRe-DPO}, a retrieval framework that directly optimizes caption generation using retrieval relevance scores. 
At its core is Dual-Group Direct Preference Optimization (DG-DPO), a novel learning strategy that supervises captioning by modeling preferences across groups of distinct video and caption pairs.
In addition, we present an MLLM-based retrieval model that incorporates role-embeddings to better distinguish between textual inputs with different functional roles, such as an auxiliary caption and a text query.
Through extensive experiments, we demonstrate that CaRe-DPO significantly enhances retrieval performance by effectively leveraging auxiliary knowledge to generate fine-grained captions for retrieval.
Code is available at \href{https://github.com/mlvlab/CaReDPO}{https://github.com/mlvlab/CaReDPO}.
\end{abstract}

\begin{figure}[!t]
    \centering
    \begin{subfigure}[h]{\linewidth}
        \includegraphics[width=\linewidth]{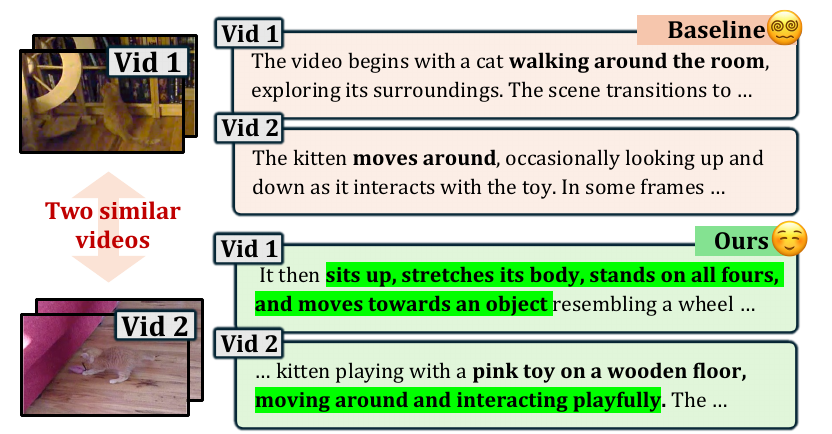}
        \caption{Generated captions for a pair of semantically similar videos}
        \label{fig:observation_1}
    \end{subfigure}
    
    \begin{subfigure}[h]{\linewidth}
        \includegraphics[width=\linewidth]{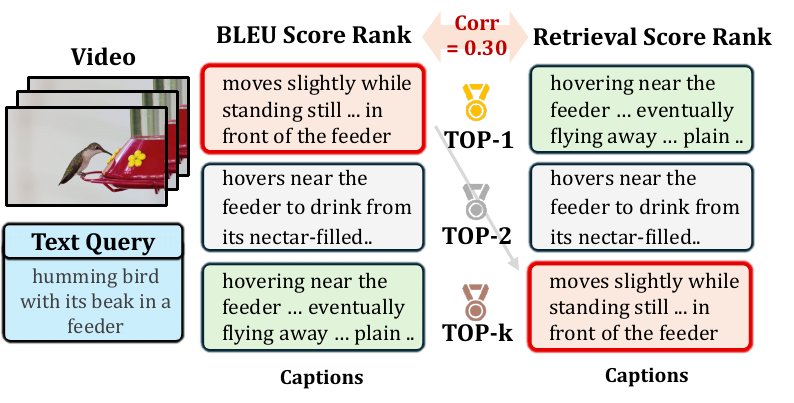}
        \caption{Comparison on rank of captions (BLEU vs Retrieval)}
        \label{fig:observation_2}
    \end{subfigure}
    
    \caption{
    \textbf{Misalignment between conventional captioners and retrieval objectives.}
    (a) Captions from pretrained models are nearly identical for similar videos, while our method produces more distinct and retrieval-relevant descriptions.
    (b) BLEU-selected top captions often mismatch retrieval rankings; correlation between the two is as low as 30\%.}
    \label{fig:observation}
    \vspace{-0.4cm}
\end{figure}
\section{Introduction}
Text-video retrieval is a fundamental task in multimodal learning, aiming to align natural language descriptions with video content.
Traditional retrieval methods often adopt dual-encoder architectures, such as CLIP~\cite{radford2021learning}, which encode videos and text queries into a shared embedding space.
However, these approaches often struggle with fine-grained semantic matching~\cite{tian2024towards, wang2023unified}, particularly when videos contain complex temporal or contextual dynamics.
To mitigate this, recent studies~\cite{wu2023cap4video, ma2024ea, hur2025narrating, yang2025expertized} have explored the use of video captions, natural language descriptions of video content, as auxiliary inputs to bridge the gap between the text queries and video content.

Multimodal Large Language Models (MLLMs) \cite{liu2024llavanext,Qwen2-VL, li2024videochat, zhang2024video, ko2023large, ko2025st, park2025deepvideo, park2024llamo} that encompass strong visual and text understandings, recently caught attention for handling multi-modal retrieval systems~\cite{lin2024mm, liu2024lamra, wei2024uniir, ko2025bidirectional}.
Their capacity to jointly attend to both visual and textual inputs allows them to interpret diverse and complex text queries in relation to video content while also leveraging captions as additional semantic context, providing a promising direction for advancing retrieval performance.

However, often captions produced by pretrained models fail to capture the detailed distinctions necessary for retrieval.
As illustrated in Fig.~\ref{fig:observation_1}, two visually similar videos depicting cats in indoor environments are given captions that are generic and overlapping, describing actions like `walking around the room' or `moves around.' 
Nevertheless, we expect to generate captions that surface discriminative visual details, such as `sits up, stretches its body' and `interacting playfully', which provide discriminative cues critical for retrieval among similar videos.
This issue is further amplified by the common practice of evaluating caption quality using metrics such as BLEU~\cite{papineni2002bleu}.
Specifically, as shown in Fig.~\ref{fig:observation_2}, among the three distinct captions generated for a video, the top-1 caption selected based on a conventional captioning metric often does not align with the top-1 caption when ranked by retrieval relevance score (placed at the bottom rank).
To quantify this misalignment, we measured the Pearson correlation between the indices of the top-1 ranked captions under each metric, finding a correlation as low as 0.30.
This highlights a substantial discrepancy between conventional captioning metric-based evaluations and retrieval-oriented objectives.

To this end, we propose \textbf{CaRe-DPO}, \textbf{Ca}ptioning for Text-Video \textbf{Re}trieval via \textbf{D}ual-Group Direct \textbf{P}reference \textbf{O}ptimization, a novel retrieval framework.
At the core is Dual-Group Direct Preference Optimization (DG-DPO), directly supervising the captioning model with the retrieval scores to align with the retrieval objective.
Unlike standard single-group DPO (local caption ranking within a video), DG-DPO incorporates dual-group preferences, learning global ranking over video-caption pairs.
In addition, we introduce role-embeddings during retrieval model training to differentiate the roles of heterogeneous textual inputs, allowing the model to more effectively leverage the auxiliary captions.
We empirically validate that CaRe-DPO encourages the MLLM-based retrieval model to further leverage the auxiliary captions during retrieval and enables enhancement of the caption quality, yielding a performance improvement across various text-video retrieval benchmarks.

The main contributions of this work are:
\begin{itemize}
\item To the best of our knowledge, we are the first to address the misalignment between conventional captioning metrics and retrieval objectives, and tackle the challenge of leveraging captions to improve retrieval performance.
\item We propose \textbf{CaRe-DPO}, a retrieval framework that integrates role-embeddings with retrieval-aligned caption optimization to leverage auxiliary captions in MLLM-based text-video retrieval.
\item Our DG-DPO supervises caption generation using retrieval relevance scores with both local (within-video) and global (cross-video-caption) ranking, enabling generation of captions that better reflect retrieval importance.
\item Our CaRe-DPO improves caption quality and retrieval performance, achieving superior performance across multiple benchmarks.
\end{itemize}

\section{Related Work}
\subsection{Text-Video Retrieval}
To improve text-video retrieval, recent studies have explored the use of captions as auxiliary supervision.
Cap4Video~\cite{wu2023cap4video} treats them as data augmentation to generate new training pairs, enhancing cross-modal interaction.
NarVid~\cite{hur2025narrating} uses frame-level captions to enrich video understanding and applies a hard negative loss for better discrimination.
ExCae~\cite{yang2025expertized} refines captions through self-learning to enhance expressiveness while minimizing manual intervention.
Recently, with the advancement of Multimodal Large Language Models (MLLMs), several works~\cite{lin2024mm, liu2024lamra, ko2025bidirectional} introduced MLLMs in multi-modal retrieval systems.
MM-Embed~\cite{lin2024mm} finetuned the MLLMs to universal retrievers, adopting thought prompt-and-reranking strategies.
Concurrently with our work, BLiM~\cite{ko2025bidirectional} investigates candidate prior bias induced by candidate likelihood estimation and improves retrieval performance through bidirectional likelihood estimation.

\subsection{Direct Preference Optimization}
Direct Preference Optimization (DPO)~\cite{rafailov2023direct} has emerged as an efficient alternative to reinforcement learning from human feedback (RLHF)~\cite{christiano2017deep,ouyang2022training,stiennon2020learning} for aligning large language models with human preferences.
Recent studies have explored several limitations of DPO.
To mitigate length bias in preference data, prior approaches introduce reward normalization~\cite{meng2024simpo}, token-level probability down-sampling~\cite{lu2024eliminating}, and explicit length regularization~\cite{park2024disentangling}.
Other studies attempt to eliminate the reliance on a reference model to reduce computational cost~\cite{meng2024simpo,xu2024contrastive,hong2024orpo}.
In multimodal, DPO has been adapted to align MLLMs for tasks such as visual question-answering~\cite{li2024multi}, dense video captioning~\cite{lee2025vidchain}, and hallucination mitigations~\cite{ouali2024clip,wang2024mdpo}.
In this work, we provide a retrieval-oriented preference modeling with MLLMs.

\section{Preliminary}

\subsection{Text-Video Retrieval}
Text-Video Retrieval consists of two tasks, video-to-text retrieval (V2T) and text-to-video retrieval (T2V), which aim to find the most relevant text or video given the query among the candidates of video or text.

Often to enhance the cross-modal retrieval, several works~\cite{wu2023cap4video, yang2025expertized, hur2025narrating} propose to utilize the generated caption $\mathbf{c}^{(i)}$ of the given video $\mathbf{v}^{(i)}$ to bridge the modality gap with the textual query $\mathbf{t}^{(i)}$.
Hence, the retrieval dataset can be defined as $\mathcal{D}_{\text{ret}} = \{\mathbf{v}^{(i)}, \mathbf{c}^{(i)}, \mathbf{t}^{(i)}\}_{i=1}^N$, where $\mathbf{c}^{(i)}$ is often sampled from a pretrained captioning model ${\mathcal{M}}_{\text{cap}}$.
During inference, $\mathbf{c}^{(i)}$ is paired with the $\mathbf{v}^{(i)}$, which T2V retrieval for instance is defined as:
\begin{equation}
\label{eq:VC2T,T2VC}
    i^*_{\text{T2V}} = \arg \max_{i} P(\mathbf{v}^{(i)}, \mathbf{c
    }^{(i)}| \mathbf{t}).
\end{equation}

Recently, MLLMs have often been employed for multi-modal retrieval systems, where they are adopted to re-rank the top-$k$ text-video candidate pairs based on joint text-video similarity.
Typically, given the video $\mathbf{v}={[v_1, .., v_{N_v}]} \in \mathbb{R}^{N_v \times D}$, caption $\mathbf{c}={[c_1, .., c_{N_c}]} \in \mathbb{R}^{N_c \times D}$, and text $\mathbf{t}={[t_1, .., t_{N_t}]} \in \mathbb{R}^{N_t \times D}$, where $N_v$, $N_c$, $N_t$, and $D$ denote the numbers of video, caption, text tokens, and the hidden dimension respectively, the objective for reranking with MLLM-based models for retrieval can be defined as follows:
\begin{equation}
\label{eq:retrieval-loss}
   \mathcal{L} = -\log P(y|\mathbf{v}, \mathbf{c}, \mathbf{t}, \mathbf{I}).
\end{equation}
The output $y$ is defined with $y\in\{\texttt{True}, \texttt{False} \}$ tokens, resembling a binary classification task, and note that the auxiliary caption $\mathbf{c}$ is simply concatenated to the video along with the text query $\mathbf{t}$.
Also, $\mathbf{I}$ denotes the instruction prompt to answer `\texttt{True}' or `\texttt{False}' that is omitted for the following notations.
Hence, for the matching triplets, \textit{i.e.}, $(\mathbf{v}^{(i)}, \mathbf{c}^{(i)}, \mathbf{t}^{(j)})$ where $i=j$, the model is trained to output `\texttt{True}', while for the unmatching triples where $i \neq j$ the model is expected to output `\texttt{False}'.
During inference, following \citet{lin2024mm} and \citet{liu2024lamra} the typical approach of measuring the relevance score $s$ is:
\begin{equation}
\label{eq: inference-standared-retrieval}
s(\mathbf{v},\mathbf{c},\mathbf{t}) = \log P(y^{+}| \mathbf{v}, \mathbf{c}, \mathbf{t}),
\end{equation}
where $y^+=\texttt{True}$.
For T2V, the retrieved video is chosen as the candidate with the highest relevance score for a given text query $\mathbf{t}^{(i)}$, and vice versa for V2T.
However, we observe that simply concatenating the caption $\mathbf{c}$ into the input hinders the model from differentiating between the heterogeneous textual inputs of the text query $\mathbf{t}$ and the auxiliary caption $\mathbf{c}$.
We also empirically observe that the simple strategy of measuring the relevance score with the probability of predicting the `\texttt{True}' lacks fine-grained sensitivity required for retrieval.
\vspace{-0.2cm}
\begin{figure*}[!ht] 
    \centering
    \includegraphics[width=1.0\linewidth]{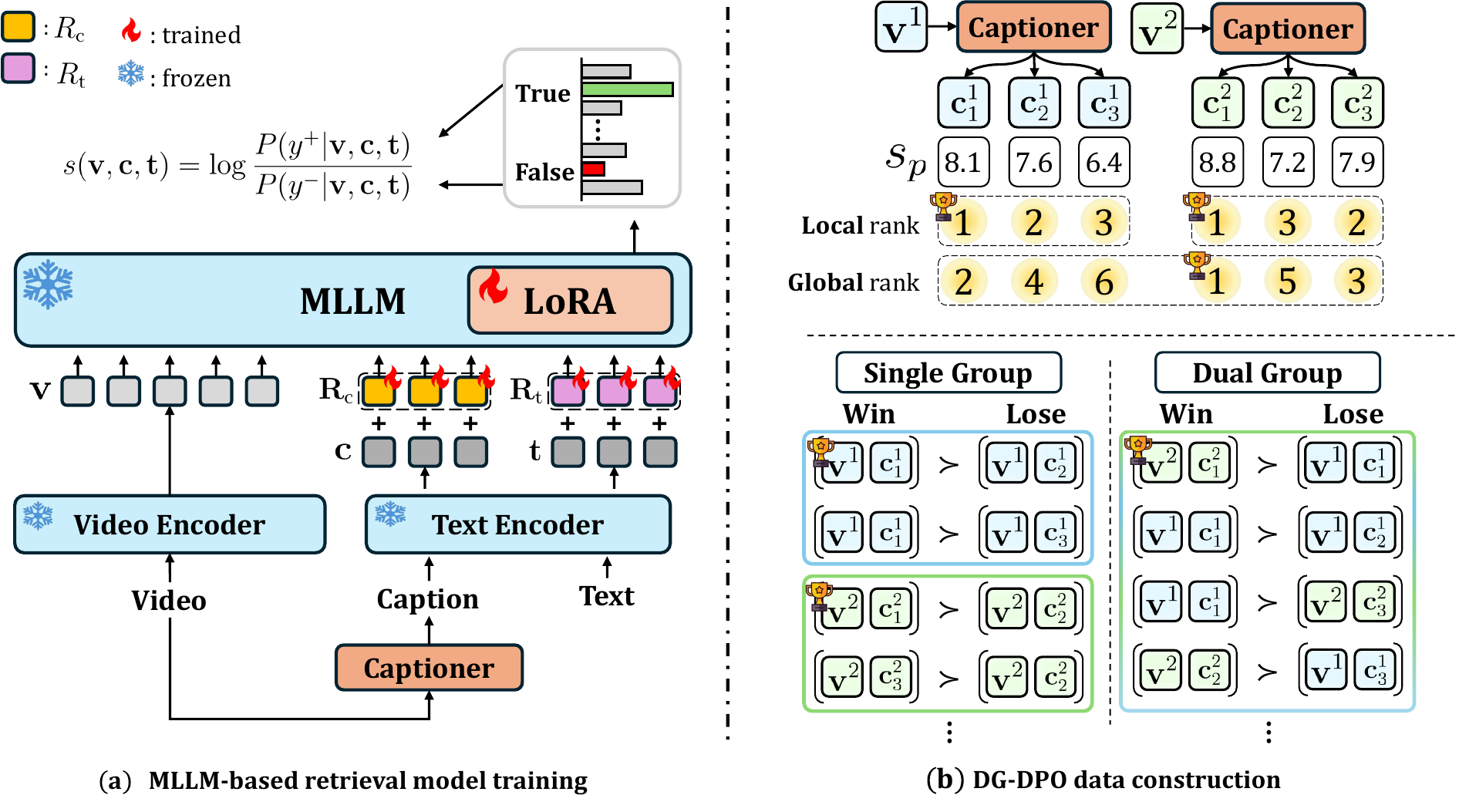}
    \caption{\textbf{Illustration of our CaRe-DPO framework.} 
    (a) depicts the MLLM-based retrieval model where we propose to adopt retrieval role-embeddings $\mathbf{R}_{\text{c}}$ and $\mathbf{R}_{\text{t}}$ for the heterogeneous textual inputs applied to each token, accordingly: auxiliary caption (orange) and retrieval target text (purple).
    In addition, we illustrate the contrastive inference strategy (contrasting the probability of generation \texttt{`True'} to \texttt{`False'}).
    (b) visualizes our DG-DPO mechanism for optimizing the captioning model, where each caption given to the video is evaluated with the retrieval relevance score $s_p$. 
    During training, unlike SG-DPO, which adopts the local rank preferences, DG-DPO adopts the global rank preference, exploring across video-caption pairs.}
    \label{fig:main}
    \vspace{-0.1cm}
\end{figure*}
\subsection{Direct Preference Optimization}
Direct Preference Optimization (DPO)~\cite{rafailov2023direct}, is a typical optimization strategy adopted to align LLMs output with human preferences, which is derived from the reinforcement learning objective in RLHF~\cite{ziegler2019fine}.
$\mathcal{D}_{\text{DPO}}$ the preference dataset for DPO  can be defined with $\{ x^{(i)}, y_w^{(i)}, y_l^{(i)}\}_{i=1}^N$, where $x$ is an input, $y_w$, $y_l$ are the preferred and dispreferred outputs, and the preference is estimated by \citet{bradley1952rank}.
Typically, the objective of DPO can be written as follows:
\begin{align}
     \mathcal{L}_\text{DPO} & (\pi_\theta ; \pi_\text{ref})  = \nonumber -\mathbb{E}_{(x, y_w, y_l) \sim \mathcal{D}_{\text{DPO}}} \\
    & \ \ \ \ \ \Bigr[  
        \log \sigma  (
            \hat r_\theta(x, y_w) - \hat r_\theta(x, y_l)
        )
    \Bigr],
\end{align}
where $\hat r_\theta(x, y) = \beta \log \frac{\pi_\theta(y \mid x)}{\pi_\text{ref}(y \mid x)}$, given $\pi_\theta$ the policy model to optimize, and $\pi_\text{ref}$ the reference model, $\beta$ is a hyperparameter for regularizing the disparity of $\pi_{\theta}$ and $\pi_{\text{ref}}$, and $\sigma$ denotes the sigmoid function.

\section{Method}
In this section, we introduce our \textbf{CaRe-DPO}, \textbf{Ca}ptioning for \textbf{Re}trieval via \textbf{D}ualGroup-Direct \textbf{P}reference \textbf{O}ptimization, a novel retrieval framework that enhances text-video retrieval with auxiliary captions.
In Sec.~\ref{sec:mllm-retrieval}, we describe our retrieval backbone built upon an MLLM, which jointly encodes video and text inputs to compute cross-modal similarity, where we also introduce retrieval role-embeddings to differentiate the heterogeneous textual inputs.
Then in Sec.~\ref{sec:dg-dpo}, we present DG-DPO, a preference optimization method that supervises the captioning model to further align with the retrieval objective.
The overall framework is illustrated in Fig.~\ref{fig:main}.  

\subsection{MLLM-based Retrieval Model}
\label{sec:mllm-retrieval}
\paragraph{Training.}
Following recent trends in retrieval, we adopt MLLMs for text-video retrieval.
However, we observe an interesting phenomenon: incorporating auxiliary captions into MLLM-based retrieval models often leads to only marginal performance gains.
In particular, even when descriptive captions are provided, replacing them with random captions yields nearly identical retrieval performance (see Sec.~\ref{supp:preference_score} and \ref{sec:supp-role-embedding} in the Appendix). 
This suggests that the model does not effectively distinguish the caption’s role as auxiliary context from the text query as the retrieval target, resulting in inefficient use of the additional information.
Hence, we adopt a simple yet effective retrieval role-embedding.
Specifically, given the input triplet $(\mathbf{v},\mathbf{c},\mathbf{t})$, we introduce a new role-embeddings $R_{\text{c}}\in  \mathbb{R}^{D}$ and $R_{\text{t}}\in \mathbb{R}^{D}$, which are combined to each corresponding tokens of $\mathbf{c}$ and $\mathbf{t}$, respectively.
Hence, the training objective of Eq.~\ref{eq:retrieval-loss} can be modified as follows:
\begin{equation}
    \mathcal{L}_{\text{ret}} = -\log P(y|\mathbf{v}, \mathbf{c}+\mathbf{R}_{\text{c}}, \mathbf{t}+\mathbf{R}_{\text{t}}),
\end{equation}
where $\mathbf{R}_{\text{c}} = \mathbf{1}_{N_c}R^{\top}_{\text{c}}$ and $\mathbf{R}_{\text{t}} = \mathbf{1}_{N_t}R^{\top}_{\text{t}}$.
Such a simple approach avoids $\mathcal{M}_{\text{ret}}$ to explicitly distinguish according to its roles: the caption as contextual knowledge and the text query as retrieval targets.
Note that unless stated, $\mathbf{c}$ and $\mathbf{t}$ corresponds to $\mathbf{c} + \mathbf{R}_{\text{c}}$ and $\mathbf{t} + \mathbf{R}_{\text{t}}$ for the following notations.

\paragraph{Inference.}
For the inference stage, we empirically observe that instead of simply adopting the probability of generating the $y^{+}$ token as the retrieval relevance score (Eq.~\ref{eq: inference-standared-retrieval}), it is more effective to use the pairwise score margin between $y^{+}$ and $y^{-}$ generation as follows:
\begin{equation}
\label{eq:inference}
    s(\mathbf{v}, \mathbf{c}, \mathbf{t})= \log  \frac{P(y^+| \mathbf{v}, \mathbf{c}, \mathbf{t})}{P(y^{-}| \mathbf{v}, \mathbf{c}, \mathbf{t})} 
\end{equation}
Such a contrastive inference strategy enables the retrieval model to be more sensitive to the subtle differences between the input and its output decision, thereby enhancing retrieval performance (refer to Table~\ref {supp:inference} in the Appendix). 
\subsection{DG-DPO}
\label{sec:dg-dpo}
\paragraph{Retrieval score-driven preference dataset.}
To further improve auxiliary captions for retrieval, which arises from the misalignment between captioning evaluation and retrieval objectives, we first construct the preference dataset that directly adopts the \textit{retrieval scores} as supervision. 
First, we sample $K$ number of captions $\{\mathbf{c}^{(i)}_k\}_{k=1}^{K}$ for each video $\mathbf{v}^{(i)}$, denoted as $\mathbf{c}_{k}^{(i)}\sim \mathcal{M}_\text{cap}(\mathbf{v}^{(i)})$ where  $\mathcal{M}_\text{cap}(\cdot)$ refers to the pretrained captioning model.
Then, we adopt the retrieval model $\mathcal{M}_\text{ret}(\cdot)$ to evaluate the quality of the sampled captions for video-text retrieval.
Specifically, we adopt the relevance score between $\mathbf{c}^{(i)}_k$ and the $\mathbf{t}^{(i)}_k$ while masking the video tokens in the attention mask ($\mathbf{v}^{*}$), which empirically showed to be effective in terms of precision than that of un-masked video tokens (see Tab.~\ref{tab:relevance-score} in the Appendix).
Formally, the relevance score for preference optimization, $s_p$, is defined as:
\begin{equation}
    \begin{split}
    s_{\text{p}}(\mathbf{v}^{(i)}, \mathbf{c}^{(i)}_{k}, \mathbf{t}^{(i)}) & = 
         \log  \frac{P (y^{+}| \mathbf{v}^{*}, \mathbf{c}^{(i)}_{k}, \mathbf{t}^{(i)})}{P(y^{-}|\mathbf{v}^{*}, \mathbf{c}^{(i)}_{k}, \mathbf{t}^{(i)})} \\
    \end{split}
\end{equation}

\paragraph{Single-Group Direct Preference Optimization.}
Similar to the conventional DPO approach, we define Single-Group Direct Preference Optimization (SG-DPO) as the variant where preference pairs are constructed by comparing outputs conditioned on a single input \textit{i.e.}, \textit{local retrieval rank preferences}, as illustrated under ‘local rank’ in Fig.~\ref{fig:main} (b).
Specifically, given a single video $\mathbf{v}^{(i)}$ with its associated two sampled captions of preferred $\mathbf{c}^{(i)}_{w}$ and dispreferred $\mathbf{c}^{(i)}_{l}$, preference pair $\mathbf{c}^{(i)}_{w} \mid \mathbf{v}^{(i)} \succ \mathbf{c}^{(i)}_{l} \mid \mathbf{v}^{(i)}$ satisfy the following condition:
\begin{equation}
    s_p\left (\mathbf{v}^{(i)}, \mathbf{c}^{(i)}_{w}, \mathbf{t}^{(i)} \right) > s_p \left(\mathbf{v}^{(i)}, \mathbf{c}^{(i)}_{l}, \mathbf{t}^{(i)} \right) + \gamma.
\end{equation}
$\gamma$ refers to the margin threshold, which enforces a minimum difference between retrieval scores.

\paragraph{Dual-Group Direct Preference Optimization.}
On the other hand, our proposed Dual-Group Direct Preference Optimization (DG-DPO) builds upon the SG-DPO, which extends the framework to consider preferences across distinct video-caption pairs by leveraging their associated retrieval relevance scores across the dataset, \textit{i.e.}, \textit{global retrieval rank} preferences.
For instance, given two video-caption pairs $( \mathbf{v}^{(i)}, \mathbf{c}^{(i)}_{k})$ and $(\mathbf{v}^{(j)}, \mathbf{c}^{(j)}_{k})$, where the former denote any $k$-th caption and video for the $i$-th sample, and the latter denote any $k$-th caption and the video for the $j$-th sample, the preference pair among the video-caption pair \textit{i.e.}, $\mathbf{c}^{(i)}_{w} | \mathbf{v}^{(i)} \succ \mathbf{c}^{(j)}_{l} | \mathbf{v}^{(j)}$, satisfy the following condition:
\begin{equation}
     s_p\left (\mathbf{v}^{(i)}, \mathbf{c}^{(i)}_{w}, \mathbf{t}^{(i)} \right) > s_p \left(\mathbf{v}^{(j)}, \mathbf{c}^{(j)}_l, \mathbf{t}^{(j)} \right ) + \gamma. 
\end{equation}
Notably, the preference pairs of DG-DPO include cases satisfying both $i =j$ and $i \neq j$, whereas SG-DPO only considers sample pairs with $i=j$.
Overall, the model learns to prefer video-caption pairs, which results in higher retrieval relevance scores, while considering both the local rank preference of the caption and the global rank preference across distinct video-caption pairs, enhancing the retrieval performance.
Hence, $\mathcal{L}_{\text{DG-DPO}}$ can be written as:
\begin{equation}
\begin{split}
\label{eq:DG-DPO}
     & \ \ \mathcal{L}_\text{DG-DPO}  = -\mathbb{E}_{(\mathbf{v}^{(i)}, \mathbf{v}^{(j)}, \mathbf{c}^{(i)}_w, \mathbf{c}^{(j)}_l) \sim \mathcal{D_{\text{DG-DPO}}}} \\ 
    & \ \ \ \left[
        \lambda_{i,j} \cdot 
        \log \sigma \left(
           \hat{r}_{\theta}(\mathbf{v}^{(i)}, \mathbf{c}_{w}^{(i)})  - 
           \hat{r}_{\theta}(\mathbf{v}^{(j)}, \mathbf{c}_{l}^{(j)}) 
        \right)
    \right].
\end{split}
\end{equation}
Note $\lambda_{i,j}$ serves as a weighting factor that balances the contribution between pairs with $i=j$ and $i\neq j$.
Importantly, this does not require additional training samples; instead, we reuse the pre-computed log probability values from the computation from when $i = j$ to compute $\mathcal{L}_{\text{DG-DPO}}$ for $i \neq j$.
Hence, we effectively leverage the samples within the same batch-aggregated across multiple GPUs to adopt the global rank of video-caption pairs.
As a result, without substantial computational or memory overhead, the captioning model is encouraged to explore consistent ranking preferences across a wider range of sample combinations of video-caption pairs for retrieval with auxiliary captions.

\begin{table*}[!ht]
    \centering
    \begin{adjustbox}{width=\textwidth}
    \setlength{\tabcolsep}{2.0pt}
    \scriptsize
   \begin{tabular}{l|c c c|c c c|c c c|c c c|c c c|c c c}
        \toprule
        & \multicolumn{9}{c|}{\textbf{Text-to-Video}} 
        & \multicolumn{9}{c}{\textbf{Video-to-Text}} \\
        & \multicolumn{3}{c|}{\textbf{DiDeMo}} 
        & \multicolumn{3}{c|}{\textbf{ActivityNet}} 
        & \multicolumn{3}{c|}{\textbf{MSRVTT}} 
        & \multicolumn{3}{c|}{\textbf{DiDeMo}} 
        & \multicolumn{3}{c|}{\textbf{ActivityNet}} 
        & \multicolumn{3}{c}{\textbf{MSRVTT}} \\
        & R@1 & R@5 & R@10  
        & R@1 & R@5 & R@10 
        & R@1 & R@5 & R@10 
        & R@1 & R@5 & R@10 
        & R@1 & R@5 & R@10 
        & R@1 & R@5 & R@10 \\
\midrule
        \midrule
        \multicolumn{19}{l}{\textbf{\textit{Non-MLLM-based}}} \\
        CLIP4Clip~\cite{luo2022clip4clip} 
            & 42.8 & 68.5 & 79.2 & 40.5 & 72.4 & 83.4 & 44.5 & 71.4 & 81.6
            & 42.5 & 70.6 & 80.2 & 42.6 & 73.4 & 85.6 & 43.1 & 70.5 & 81.2 \\
        ViCLIP~\cite{wang2024internvid} 
            & 49.4 & - & - & 49.8 & - & - & 52.5 & - & - 
            & 50.2 & - & - & 48.1 & - & - & 51.8 & - & - \\  
        MV-Adapter~\cite{jin2024mv} 
            & 44.3 & 72.1 & 80.5 & 42.9 & 74.5 & 85.7 & 46.2 & 73.2 & 82.7
            & 42.7 & 73.0 & 81.9 & 43.6 & 75.0 & 86.5 & 47.2 & 74.8 & 83.9 \\
        InternVideo~\cite{wang2022internvideo} 
            & 57.9 & 82.4 & 88.9 & 62.2 & 85.9 & 93.2 & 55.2 & 79.6 & 87.5
            & 59.1 & 81.8 & 89.0 & 62.8 & 86.2 & 93.3 & 57.9 & 79.2 & 86.4 \\
        UMT~\cite{li2023unmasked} 
            & 70.4 & 90.1 & 93.5 & 66.8 & 89.1 & 94.9 & 58.8 & 81.0 & 87.1
            & 67.9 & 88.6 & 93.0 & 64.4 & 89.1 & 94.8 & 58.6 & 81.6 & 86.5 \\
        Cap4Video~\citep{wu2023cap4video} 
            & 52.0 & 79.4 & 87.5 & - & - & - & 51.4 & 75.7 & 83.9
            & - & - & - & - &  - & - & 49.0 & 75.2 & 85.0 \\
        NarVid~\citep{hur2025narrating} 
        & 53.4 & 79.1 & 86.3 
        & - & - & - 
        & 52.7 & 77.7 & 85.6 
        
        & - & - & - 
        & - & - & - 
        & 51.1 & 76.8 & 85.2 \\
        InternVideo2 1B$^*$~\cite{wang2024internvideo2} 
            & 75.3 & 92.5 & 95.8 & 68.8 & 89.7 & 94.7 & 59.4 & 80.9 & 86.6
            & 73.1 & 92.1 & 94.9 & 65.3 & 88.0 & 94.2 & 56.9 & 76.9 & 84.6 \\
        InternVideo2 6B~\cite{wang2024internvideo2} 
            & 74.2 & - & - & 74.1 & - & - & 62.8 & - & -
            & 71.9 & - & - & 68.7 & - & - & 60.2 & - & - \\
        \midrule
        \midrule
        \multicolumn{19}{l}{\textbf{\textit{MLLM-based$\dagger$}}} \\
        MM-Embed~\cite{lin2024mm} 
            & 81.6 & 94.9 & \textbf{96.3} & 78.5 & - & - & 61.2 & 82.7 & \textbf{88.8}
            & 79.7 & 94.9 & 96.2 & 70.7 & - & - & 60.5 & 82.3 & 87.1 \\
        LamRA~\cite{liu2024lamra} 
            & 83.5 & 94.8 & 96.2 & 76.0 & 92.8 & 96.3 & 59.7 & 81.4 & 87.2
            & 79.4 & 94.8 & \textbf{96.6} & 68.7 & 90.1 & 95.3 & 60.7 & 82.3 & \textbf{89.0} \\
        \midrule
        \rowcolor{lightblue}
        CaRe-DPO (Ours) 
            & \textbf{85.1} & \textbf{95.0} & 96.2 
            & \textbf{79.2} & \textbf{93.6} & \textbf{96.5} 
            & \textbf{64.1} & \textbf{83.8} & \textbf{88.8} 
            & \textbf{82.5} & \textbf{95.2} & 96.3 
            & \textbf{74.4} & \textbf{92.4} & \textbf{96.3} 
            & \textbf{63.8} & \textbf{83.0} & 87.3 \\
        \bottomrule
    \end{tabular}
    \end{adjustbox}
    \caption{\textbf{Comparison with state-of-the-art Text-Video Retrieval models.}
    * denotes reproduced results. 
    We also report the performance of MLLM-retrieval models, which we reproduced adequately for Text-Video Retrieval, adopting their approach while applying to the same baseline as ours, VideoChat-Flash, denoted with the $\dagger$.} 
    \label{tab:main}
    \vspace{-0.2cm}
\end{table*}

\section{Experiments}
\subsection{Experiments Setup}
\paragraph{Datasets and metrics.}
To validate the effectiveness of CaRe-DPO, we evaluate on three Text-Video retrieval benchmarks: DiDeMo~\cite{anne2017localizing}, ActivityNet~\cite{caba2015activitynet}, and MSRVTT~\cite{xu2016msr}.
More explanation of the datasets is in Sec.~\ref{app:dataset-detail} of the Appendix.
For evaluation, we adopt the standard retrieval metrics: Recall@K$\in {\{1, 5, 10\}}$.
Note that for auxiliary captions, we sample two per instance and average the performance over those to mitigate the caption variability while providing more robust results.
Unless otherwise specified, all experiments are conducted on DiDeMo, which serves as our primary benchmark. 
For additional validation, we also report results on the multi-text text-to-video retrieval benchmark MSVD~\cite{chen2011collecting} in Sec.~\ref{sec-supp:msvd} of the Appendix, which further demonstrates the effectiveness of our CaRe-DPO. 

\paragraph{Implementation details.}
For retrieval, we adopt InternVideo2-1B~\cite{wang2024internvideo2} to initially compute the similarity between the video and the text query, and then we retrieve the top-16 candidates for re-ranking.
Our MLLM-based retrieval model, capable of leveraging auxiliary captions, is built upon VideoChat-Flash-7B~\cite{li2024videochat}.
For captioning, we adopt pretrained LLaVA-OneVision-7B~\cite{li2024llava}, where we utilize 16 frames per video for all datasets, and further align the model with our DG-DPO.
For efficiency, we apply LoRA~\cite{hu2021lora} for parameter-efficient fine-tuning for both training.
See Sec.~\ref{supp:Implementation} of Appendix for more details, including the training hyperparameters.

\begin{table}[!t]
    \centering
    \begin{adjustbox}{width=\linewidth}
    \begin{tabular}{l|c c|c c|c c|c}
        \toprule
         & \multicolumn{2}{c|}{\textbf{DiDeMo}} 
         & \multicolumn{2}{c|}{\textbf{ActivityNet}} 
         & \multicolumn{2}{c|}{\textbf{MSRVTT}} 
         & \multirow{2}{*}{\shortstack{$\text{Avg.}$ \\ $\Delta$}} \\
         & T2V & V2T & T2V & V2T & T2V & V2T & \\
        \midrule
        Baseline & 83.1 & 79.6 & 78.3 & 74.0 & 62.7 & 63.6 & - \\
        + $\mathcal{L}_{\text{SFT}}$ & 82.6 & 82.0 & 78.0 & 73.9 & 62.9 & 63.0 & (+0.2) \\
        \rowcolor{lightblue}
        + $\mathcal{L}_{\text{SG-DPO}}$ & 84.4 & 82.4 & 78.9 & 74.1 & 63.4 & 63.3 & (+0.9) \\
        \rowcolor{lightblue}
        + $\mathcal{L}_{\text{DG-DPO}}$ & \textbf{85.1} & \textbf{82.5} & \textbf{79.2} & \textbf{74.4} & \textbf{64.1} & \textbf{63.8} & (+1.3) \\
        \bottomrule
    \end{tabular}
    \end{adjustbox}
    \caption{\textbf{Ablation on training objectives for $\mathcal{M}_{_\text{cap}}$.}
    R@1 retrieval performance from different training objectives for $\mathcal{M}_{\text{cap}}$. 
    `Avg. $\Delta$' denotes the R@1 point increase compared to the baseline across datasets.}
    \label{tab:ablation-db-dpo}
    \vspace{-0.2cm}
\end{table}

\subsection{Experimental Results}
\paragraph{Main results.}
Tab.~\ref{tab:main} shows the performance of the State-of-the-Art text-video retrieval models, including non-MLLM-based and MLLM-based.
The results show that our CaRe-DPO outperforms baseline models across various datasets, especially in R@1 for both T2V and V2T.
Among non-MLLM-based models, ours effectively improves performance over the SOTA model of InternVideo2-6B, with an average percentage increase of 14.7\%,7.7\%, and 4.1\% in R@1 for DiDeMo, ActivityNet, and MSRVTT, respectively.
To further validate the effectiveness of our framework across MLLM-based retrieval models, we compare against MM-Embed and LamRA.
Notably, our CaRe-DPO shows superior performance with an average percentage increase in R@1 of 3.9\%, 3.1\%, and 5.1\% compared to MM-Embed, and 2.9\%, 6.3\%, and 6.2\% compared to LamRA across datasets. 
Overall, CaRe-DPO consistently outperforms the baselines, highlighting its effectiveness in enhancing text-video retrieval with MLLM-based models.

\begin{table}[!t]
    \centering
    \begin{adjustbox}{width=1.0\linewidth}
    \begin{tabular}{c|c|c c|c c|c c}
        \toprule
          & &\multicolumn{2}{c|}{\textbf{DiDeMo}} 
          & \multicolumn{2}{c|}{\textbf{ActivityNet}} 
          & \multicolumn{2}{c}{\textbf{MSRVTT}} \\
        Inference & $\mathcal{M}_{\text{cap}}$  & T2V & V2T & T2V & V2T & T2V & V2T \\
        \midrule
        \multirow{2}{*}{$ s(\mathbf{c}, \mathbf{t})$} 
            & Baseline  & 49.6 & 40.8  & 43.2 & 37.0 & 40.5 & 37.7 \\
            & \cellcolor{lightblue}  + $\mathcal{L}_{\text{DG-DPO}}$ 
            & \cellcolor{lightblue} 51.2 & \cellcolor{lightblue} 43.4  
            & \cellcolor{lightblue} 53.0 & \cellcolor{lightblue} 43.6 
            & \cellcolor{lightblue} 49.1 & \cellcolor{lightblue} 45.5 \\
        \midrule
        \multirow{2}{*}{$ s(\mathbf{v}, \mathbf{c})$} 
            & Baseline  & 91.8 & 90.7 & 88.2 & 86.5 & 88.9 & 86.1 \\
            & \cellcolor{lightblue} + $\mathcal{L}_{\text{DG-DPO}}$ 
            & \cellcolor{lightblue} 92.1 & \cellcolor{lightblue} 92.2 
            & \cellcolor{lightblue} 88.7 & \cellcolor{lightblue} 87.5  
            & \cellcolor{lightblue} 89.7 & \cellcolor{lightblue} 87.1 \\
        \bottomrule
    \end{tabular}
    \end{adjustbox}
    \caption{\textbf{Analysis on caption quality for retrieval.}
    `Baseline' denotes zero-shot captions adopted for retrieval.
    For $s(\mathbf{c}, \mathbf{t})$, we adopt the model trained with $(\mathbf{v}, \mathbf{c}, \mathbf{t})$ while masking the video tokens.
    For $s(\mathbf{v}, \mathbf{c})$, we utilize the model trained solely on $(\mathbf{v}, \mathbf{t})$.
    We report R@1 performance for both T2V and V2T.}
    \label{tab:ablation-v2c}
\end{table}
\begin{table}[!t]
    \centering
    \renewcommand{\arraystretch}{1.2}
    \setlength{\tabcolsep}{2pt}
    \begin{adjustbox}{width=1.0\linewidth}
    \begin{tabular}{c|c|c|ccc|ccc|c}
        \toprule
        \textbf{Train Input} & \multirow{2}{*}{\textbf{R$_{\text{c}}$}, \textbf{R$_{\text{t}}$}} & \textbf{Inf.}  & \multicolumn{3}{c|}{\textbf{Text-to-Video}} & \multicolumn{3}{c|}{\textbf{Video-to-Text}} & \multirow{2}{*}{\shortstack{$\text{Avg.}$ \\ $\Delta$}}\\
        $\mathcal{L}_{\text{ret}}(\cdot)$ & & cap. $\mathbf{c}$  & R@1 & R@5 & R@10  & R@1 & R@5 & R@10 & \\
        \midrule
        \midrule
        \multirow{2}{*}{$(\mathbf{v}, \mathbf{c}, \mathbf{t})$} 
            & \ding{55} & rand. & 81.5 & 94.6 & 95.9 & 79.1 & 94.6 & 96.5 & - \\
            & \ding{55} &  orig. & 81.6 & 94.3 & 95.9 & 79.2 & 94.7 & 96.7 & (+0.1) \\
        \midrule
        \multirow{2}{*}{$(\mathbf{v}, \mathbf{c}', \mathbf{t}')$} 
            & \ding{52} & rand. & 82.6 & 94.4 & 96.0 & 76.5 & 95.0 & 96.2 & - \\
            & \cellcolor{lightblue} \ding{52} & \cellcolor{lightblue} orig.   & \cellcolor{lightblue} 83.1 & \cellcolor{lightblue} 94.4 & \cellcolor{lightblue} 96.2 & \cellcolor{lightblue} 79.6 & \cellcolor{lightblue} 94.6 & \cellcolor{lightblue} 96.6 & \cellcolor{lightblue} (+1.8) \\
        \bottomrule
    \end{tabular}
    \end{adjustbox}
    \caption{\textbf{Ablation on the role-embeddings of $\mathcal{M}_{\text{ret}}$.}
    We adopt the zero-shot captions with the standard inference strategy.
    `rand' and `orig.' denote \textit{random} and \textit{original} captions, respectively, and `Inf.' denotes the inference stage.
    $\mathbf{c}'$ and $\mathbf{t}'$ denote $\mathbf{c}+\mathbf{R}_{\text{c}}$ and $\mathbf{t}+\mathbf{R}_{\text{t}}$, respectively. 
    `Avg. $\Delta$' denotes an average change in R@k performance, between `rand' and `orig' captions.
    }
    \label{tab:ablation-role-embeddings}
    \vspace{-0.1cm}
\end{table}

\subsection{Quantitative Analysis}
\paragraph{Ablation on training objectives for $\mathcal{M}_{\text{cap}}$.}
In Tab.~\ref{tab:ablation-db-dpo}, we analyze different objectives for training the captioning model on the performance of text-video retrieval.
As shown, simply fine-tuning on the given dataset, denoted as $\mathcal{L}_{\text{SFT}}$ (row 2), results in an average R@1 improvement of 0.2 points, with a per-dataset increase of 1.0 for DiDeMo and a decrease of 0.2 for both ActivityNet and MSRVTT.
In contrast, adopting our $\mathcal{L}_{\text{SG-DPO}}$  or $\mathcal{L}_{\text{DG-DPO}}$, which optimizes the model with DPO using retrieval scores for preference determination, results in superior performance.
Specifically, $\mathcal{L}_{\text{SG-DPO}}$ (row 3), which relies on the local preference of the retrieval score, shows point increases of 2.1, 0.3, and 0.2 for DiDeMo, ActivityNet, and MSRVTT, respectively.
By further considering the global preference based on retrieval scores, $\mathcal{L}_{\text{DG-DPO}}$ (row 4) achieves even higher retrieval precision, with point increases of 2.5, 0.8, and 0.8 compared to the baseline across the datasets.
These results highlight the effectiveness of leveraging retrieval scores with DPO to better align generated captions for retrieval, and demonstrate that DG-DPO, which accounts for global preferences beyond local video-caption pairs, further improves performance.

\begin{figure*}[!ht] 
    \centering
    \includegraphics[width=\linewidth]{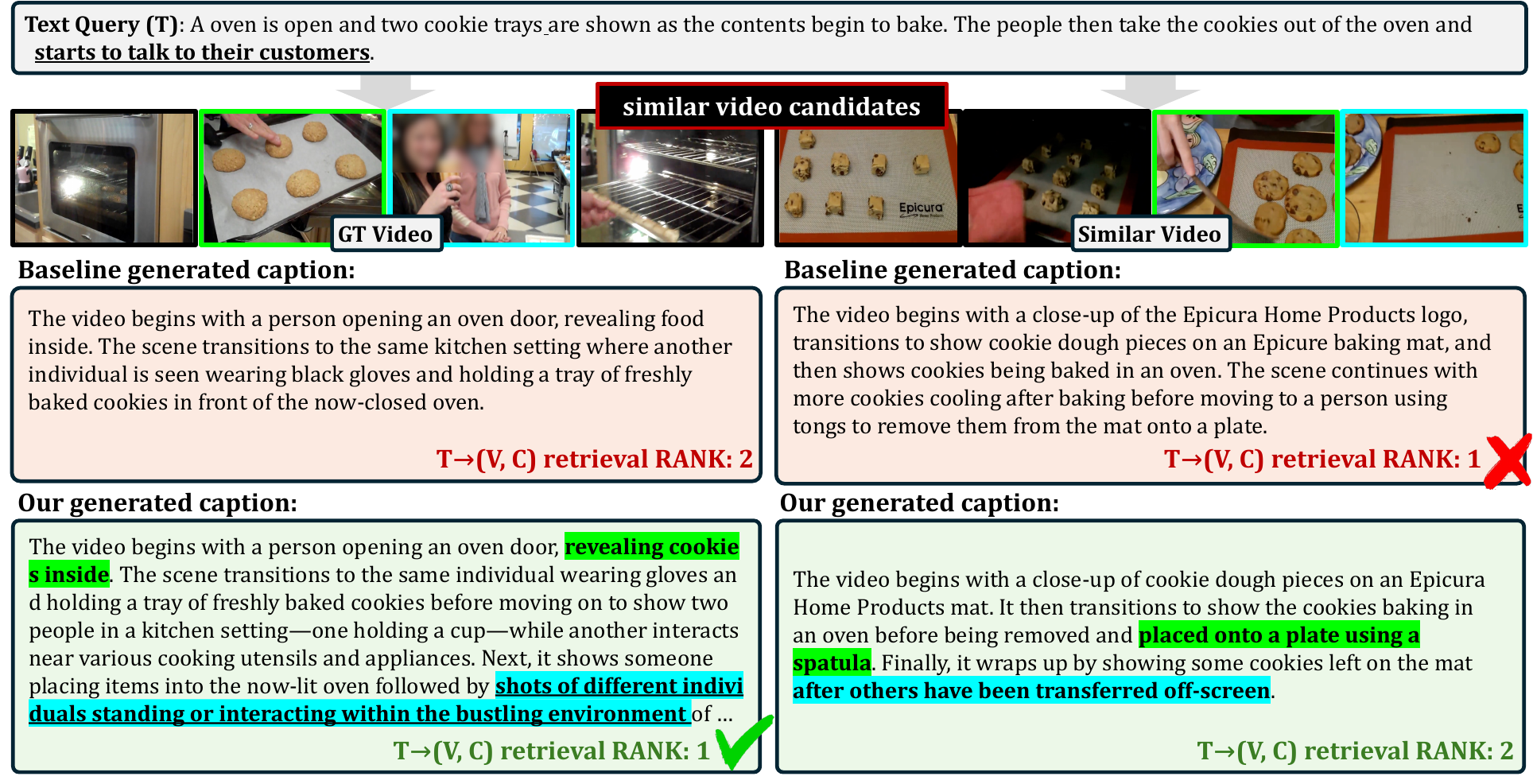}
    \caption{\textbf{Qualitative example on the effect of generated video caption.}
    Comparison of captions generated for visually similar videos on ActivityNet (left: ground-truth video, right: highly similar incorrect video) using the zero-shot captioning model (Baseline) and our DG-DPO-optimized model (Ours).
    Note the fine-grained details captured by our model but omitted in the baseline are highlighted in color.
    Additionally, the border color of each video frame corresponds to the caption depicted in our generated caption, and the underline denotes the fine-grained details that closely match the input text query (T).
    T $\rightarrow$ (V, C) denotes text-to-video retrieval, where C is the auxiliary caption for the given video.
    We also show the retrieval rank for each candidate with different generated captions. 
    The qualitative results demonstrate that the DG-DPO-optimized caption improves retrieval, increasing the ground-truth video’s rank from 2 to 1 in T2V.
}
    \label{fig:qual}
    \vspace{-0.3cm}
\end{figure*}

\paragraph{Analysis on the quality of caption for retrieval.}
To further investigate the effectiveness of captions in retrieval with CaRe-DPO, we design a series of experiments shown in Tab.~\ref{tab:ablation-v2c}: text-to-caption (T2C) (upper half) and video-to-caption retrieval (V2C) (lower half).
T2C assesses how well the auxiliary caption semantically aligns with the query, V2C measures the degree to which the caption captures the distinctive content of the video itself. 
The results show that the captions generated from $\mathcal{M}_{\text{cap}}$ trained with $\mathcal{L}_{\text{DG-DPO}}$ result in consistent improvements across both retrieval tasks.
Specifically, in T2C, the caption generated after adopting our DG-DPO yields an average increase of 7.6 points, especially in MSRVTT, with an 8.6 points increase in T2V and a 7.8 points increase in V2T.
Also notable in ActivityNet with a 9.8 points and a 6.6 points increase in T2V and V2T, respectively.
In V2C, the zero-shot caption itself shows strong explanability of the video, yet with our $\mathcal{L}_{\text{DG-DPO}}$, it further leads to performance enhancement with an average 0.9, 0.8, and 0.9 points increase in R@1 on DiDeMo, ActivityNet, and MSRVTT, respectively.

\paragraph{Effectiveness of retrieval role-embeddings.}
Tab.~\ref{tab:ablation-role-embeddings} presents the impact of retrieval role-embeddings for the MLLM-based retrieval model.
As shown, when replacing the caption with a random caption, the model shows a minimal performance drop of 0.1 in R@1 on average (compare rows 1 and 2).
In contrast, using the same caption, training with role-embeddings results in a superior performance with 83.1 for T2V and 79.6 in V2T (compare rows 2 and 4) while showing higher sensitivity to the quality of the caption with a notable 1.8 improvement in average for R@1 compared to the one with a random caption (compare rows 3 and 4).
These results highlight the effectiveness of role-embeddings in differentiating the two roles of auxiliary knowledge and retrieval target, leading to more accurate retrievals.
Note that we use standard pretrained captions in this ablation to isolate the effect of retrieval role-embeddings; further experiments with DG-DPO-optimized captions are provided in Tab.~\ref{supp-tab:ablation-role-embeddings-finetuned} of the Appendix.

\begin{table}[!t]
    \centering
    \begin{adjustbox}{width=1.0\linewidth}
    \begin{tabular}{c|c|c c|c}
        \toprule
        \textbf{Preference score} & 
        $\mathcal{M}_{\text{cap}}$ & \textbf{T2V} & \textbf{V2T} & 
        \textbf{Avg. Confidence} \\
        \midrule
        \midrule
        \multirow{2}{*}{BLEU} 
            & + $\mathcal{L}_{\text{SG-DPO}}$ & 82.9 & 74.2 & 82.7 \\
            & + $\mathcal{L}_{\text{DG-DPO}}$ & 83.7 & 75.4 & 83.1 \\
        \midrule
        \multirow{2}{*}{Retrieval Score} 
            & + $\mathcal{L}_{\text{SG-DPO}}$ & 83.7 & 75.4 & 86.9  \\
            & + $\mathcal{L}_{\text{DG-DPO}}$ & \textbf{84.5} & \textbf{77.4} & \textbf{89.6}  \\
        \bottomrule
    \end{tabular}
    \end{adjustbox}
    \caption{\textbf{Effectiveness of DG-DPO in challenging retrieval cases.} 
    Results of T2V and V2T R@1 in challenging retrieval cases with highly similar video candidates (pairwise average cosine similarity averaged over frames $>$ 0.97).}
    \label{tab:hard-retrieval}
    \vspace{-0.2cm}
\end{table}
\begin{table}[ht!]
    \centering
    \begin{adjustbox}{width=1.0\linewidth}
    \begin{tabular}{l|c c c}
        \toprule
        \textbf{Training Strategy} & \textbf{Self-BLEU ($\downarrow$)} & \textbf{Distinct-1 ($\uparrow$)} & \textbf{Distinct-2 ($\uparrow$)} \\
        \midrule
        \midrule
        Zero-shot & 0.50 & 0.08 & 0.41 \\
        + $\mathcal{L}_{\text{SFT}}$ & 0.52 & 0.08 & 0.39 \\
        + $\mathcal{L}_{\text{DG-DPO}}$ & \textbf{0.47} & \textbf{0.10} & \textbf{0.43} \\
        \bottomrule
    \end{tabular}
    \end{adjustbox}
    \caption{\textbf{Caption diversity with different training strategies.} DG-DPO consistently improves diversity and reduces redundancy in caption generation.}
    \label{tab:diversity}
    \vspace{-0.5cm}
\end{table}

\paragraph{Effectiveness of DG-DPO in challenging retrieval cases.}
Tab.~\ref{tab:hard-retrieval} highlights the effectiveness of DG-DPO in retrieval scenarios involving highly similar video candidates, from which we left samples from the test set where any pairwise video cosine similarity (averaged over frames) is greater than 0.97.
First, when comparing BLEU and retrieval score as the preference signal under SG-DPO, the retrieval score consistently yields higher average model confidence measured with Eq.~\ref{eq:inference} and scaled to a percentage (out of 100) for interpretability (86.9 vs. 82.7), indicating its stronger alignment with retrieval objectives. 
More importantly, DG-DPO further improves performance and model confidence over SG-DPO across both preference signals. 
For instance, using retrieval score supervision, DG-DPO boosts T2V and V2T R@1 from 83.7 to 84.5 and 75.4 to 77.4, respectively, and increases average confidence from 86.9 to 89.6.
These results demonstrate that modeling group-level preferences across distinct video-caption pairs allows for providing more discriminative learning signals, leading to more accurate and confident retrieval in challenging cases.
\vspace{-0.1cm}

\paragraph{Effect on caption diversity.}
Tab.~\ref{tab:diversity} shows the impact of different training strategies on caption diversity measured with three different metrics of Self-BLEU~\cite{zhu2018texygen} that measure the redundancy across generated captions, and Distinct-1 and 2~\cite{li2015diversity} that capture the ratio of unique unigrams and bigrams.
While simple fine-tuning on the dataset (SFT) increases redundancy (0.52 vs. 0.50 in Self-BLEU compared to the zero-shot), likely due to overfitting to common patterns in the training data, our DG-DPO leads to significantly more diverse captions. 
It achieves the lowest Self-BLEU of 0.47 and the highest Distinct-1 of 0.10 and Distinct-2 of 0.43, indicating richer and less repetitive outputs. 
This suggests that DG-DPO not only improves discriminative supervision but also encourages the model to generate more detailed and distinctive descriptions that are essential for visually similar videos.

\subsection{Qualitative Results}
Fig.~\ref{fig:qual} presents a comparison between captions generated by a zero-shot captioning model (baseline) and our DG-DPO-trained model (ours) for two visually similar videos depicting cookie baking.
As illustrated, the baseline captions provide general scene descriptions, whereas our model generates more detailed and context-specific captions that highlight key visual cues such as `revealing cookies inside,' `placed onto a plate using a spatula,' and `individuals standing or interacting within the bustling environment.'
For the given query, the retrieval model initially ranked a visually similar but incorrect video higher (left video in Fig.~\ref{fig:qual}), which lacked the scene where `people talk to their customers’.
However, after substituting the caption with one generated by ours, the retrieval model correctly retrieved the ground-truth video, guided by the discriminative details in the caption that closely match the text query (underlined in the figure).
This demonstrates that the fine-grained captions generated by the DG-DPO optimized captioning model facilitate better differentiation between similar content, leading to improved retrieval performance.
More qualitative results in Sec.~\ref{supp:qualitatives} of the Appendix.
\section{Conclusion}

We present CaRe-DPO, a novel retrieval framework that enhances text-video retrieval with auxiliary captions. 
Our role-embeddings enable retrieval models to explicitly distinguish the roles of heterogeneous textual inputs. 
Furthermore, our Dual-Group Direct Preference Optimization aligns caption generation with retrieval relevance scores while leveraging both local and global ranks.
Through extensive experiments, we demonstrate that CaRe-DPO enhances overall retrieval accuracy across benchmarks.

\section*{Limitations}
In this work, we propose CaRe-DPO that relies on the MLLM-based models for text-video retrieval.
CaRe-DPO builds upon MLLM-based retrieval models, which inherently rely on the pre-trained multimodal knowledge encoded in the MLLM, which also includes the captioning model adopted.
As a result, the performance of our approach may be constrained by the underlying capabilities and biases of the base MLLM, especially in domain-specific or low-resource settings.
Moreover, unlike simply training the retrieval model, ours requires training both the retrieval and captioning models and generating multiple captions for DPO training, which increases overall training time, yet results in improved retrieval performance.

\section*{Acknowledgments}
This work was partly supported by the National Research Foundation of Korea (NRF) grant funded by the Korea government (MSIT) (NRF-2023R1A2C2005373), and Hanwha Vision. 

\bibliography{acl_latex}

\newpage
\appendix

\section{Dataset Details}
\label{app:dataset-detail}

\paragraph{DiDeMo.}
DiDeMo~\cite{anne2017localizing} is a text-video retrieval benchmark, namely the Distinct Describable Moments, which comprises 10K videos, which are segmented into 5-second clips for annotation, totaling 26K annotated moments. 
Each moment is richly described with references to camera movement, temporal transitions, and actions.
We treat the retrieval task as a paragraph-to-video retrieval where we concatenate all the captions within the video, following prior works~\cite{luo2022clip4clip, wu2023cap4video, li2023unmasked, wang2024internvideo2, cheng2023vindlu, hur2025narrating}.
Note that the dataset provides 8,394 training and 1,003 test samples.

\paragraph{ActivityNet.}
Activitynet~\cite{caba2015activitynet} is a text-video retrieval benchmark that is based on 19K YouTube videos, categorized into 200 activity classes.
For each class, there exists an average of 137 videos, and each video contains about 1.41 temporal activities.
Similar to DiDeMo, we aggregate all the captions per video and implement the task as a paragraph-to-video retrieval, while we evaluate on the val1 split following ~\citet{luo2022clip4clip, li2023unmasked, wang2024internvideo2, cheng2023vindlu, hur2025narrating}.

\paragraph{MSRVTT.}
The MSRVTT~\cite{xu2016msr} dataset, namely Microsoft Research Video to Text, contains 10k video clips that span across 20 categories, of which each clip is annotated by 20 sentences.
Following previous protocols~\cite{luo2022clip4clip, wang2024internvideo2, li2023unmasked, cheng2023vindlu, hur2025narrating}, we use the 9k sample set for training (which is about 180k caption-video pairs), and adopt the 1,000 clips for testing.

\paragraph{MSVD.}
The MSVD~\cite{chen2011collecting} dataset consists of 2k videos, of which each video is annotated with around 40 captions. 
Unlike previously mentioned, MSVD is a multi-text text-video retrieval benchmark where it treats each sentence as an independent sample.
Specifically, it evaluates with one-to-many ground-truth text for video-to-text retrieval.
Following previous protocols~\cite{luo2022clip4clip, wang2024internvideo2, li2023unmasked, cheng2023vindlu, hur2025narrating}, we use 1k videos for training, and 670 videos for testing.
\section{Multi-Text Text-Video Retrieval}
\label{sec-supp:msvd}
\begin{table}[!htbp]
    \centering
    \begin{adjustbox}{width=1.0\linewidth}
    \begin{tabular}{l|c|c}
        \toprule
        Method & T2V R@1 & V2T R@1 \\
        \midrule
        \multicolumn{3}{l}{\textbf{\textit{Non-MLLM-based}}} \\
        MV-Adapter~\cite{jin2024mv} & 49.4 & 71.8 \\
        NarVid~\cite{hur2025narrating} & 53.1 & - \\
        InternVideo~\cite{wang2022internvideo} & 58.4 & 76.3 \\
        UMT~\cite{li2023unmasked} & 58.2 & 82.4 \\
        InternVideo2 1B$^*$~\cite{wang2024internvideo2} & 59.0 & \underline{\textbf{85.5}} \\
        InternVideo2 6B~\cite{wang2024internvideo2} & \textbf{61.4} & 85.2 \\
        \midrule
        \multicolumn{3}{l}{\textbf{\textit{MLLM-based$\dagger$}}} \\
        MM-Embed~\cite{lin2024mm} & 59.5 & \underline{\textbf{85.5}} \\
        LamRA~\cite{liu2024lamra} & 59.0 & \underline{\textbf{85.5}} \\
        \midrule
        \rowcolor{lightblue}
        CaRe-DPO (Ours) & \underline{\textbf{59.8}} & \textbf{85.9} \\
        \bottomrule
    \end{tabular}
    \end{adjustbox}
    \caption{\textbf{Comparison with SOTA retrieval models on MSVD.} 
    We report R@1 for both T2V and V2T retrieval. 
    Note that best values are \textbf{bold}, second-best are \textbf{\underline{bold and underlined}}.
    $\dagger$ denotes reproduced on the same baseline as ours, VideoChat-Flash.}
    \label{tab:supp-msvd_r1}
\end{table}
\begin{table}[!h]
    \centering
    \begin{adjustbox}{width=0.45\linewidth}
    \begin{tabular}{l|c c}
        \toprule
         & T2V & V2T \\
        \midrule
        Baseline & 59.0 & 85.5 \\
        + $\mathcal{L}_{\text{SFT}}$ & 59.7 & 85.5 \\
        \rowcolor{lightblue}
        + $\mathcal{L}_{\text{SG-DPO}}$ & 59.7 & 85.6 \\
        \rowcolor{lightblue}
        + $\mathcal{L}_{\text{DG-DPO}}$ & \textbf{59.8} & \textbf{85.9} \\
        \bottomrule
    \end{tabular}
    \end{adjustbox}
    \caption{\textbf{Ablation on training objectives for $\mathcal{M}_{\text{cap}}$ on MSVD.} 
    R@1 retrieval performance for T2V and V2T.}
    \label{tab:ablation-msvd-a}
\end{table}

\begin{table}[!t]
    \centering
    \begin{adjustbox}{width=0.45\linewidth}
    \begin{tabular}{l|c c}
        \toprule
        $s(\mathbf{c}, \mathbf{t})$ & T2V & V2T \\
        \midrule
         Baseline & 49.5 & 76.6 \\
         + $\mathcal{L}_{\text{DG-DPO}}$ & \textbf{51.7} & \textbf{79.0} \\
        \bottomrule
    \end{tabular}
    \end{adjustbox}
    \caption{\textbf{Analysis on caption quality for retrieval on MSVD.} 
    We report R@1 for T2V and V2T.}
    \label{tab:ablation-msvd-b}
\end{table}

\paragraph{Comparison with SOTA models on MSVD.}
Tab.~\ref{tab:supp-msvd_r1} presents a comparison of state-of-the-art text-to-video (T2V) and video-to-text (V2T) retrieval performance on the MSVD, a multi-text text-video retrieval benchmark, measured using R@1.
When adopting our CaRe-DPO, the model achieves strong performance, especially in V2T, attaining 85.9, outperforming other MLLM-based approaches as well as non-MLLM-based models.

\paragraph{Ablation on training objectives on MSVD.}
Tab.~\ref{tab:ablation-msvd-a} presents an ablation study on the impact of different training objectives for $\mathcal{M}_{\text{cap}}$ on MSVD. 
We observe that incorporating the supervised fine-tuning loss ($\mathcal{L}_{\text{SFT}}$) yields a slight improvement for T2V retrieval while V2T remains unchanged. 
Introducing our ($\mathcal{L}_{\text{SG-DPO}}$ and $\mathcal{L}_{\text{DG-DPO}}$) further enhances performance, with $\mathcal{L}_{\text{DG-DPO}}$ achieving the best results of 59.8 for T2V and 85.9 for V2T.

\paragraph{Analysis on caption quality on MSVD.}
Tab.~\ref{tab:ablation-msvd-b} analyzes the effect of caption quality on retrieval performance using the text-caption retrieval $s(\mathbf{c}, \mathbf{t})$, measuring how well the auxiliary caption semantically aligns with the query.
Compared to the baseline, adding $\mathcal{L}_{\text{DG-DPO}}$ significantly improves R@1 metrics for both T2V and V2T, reaching 51.7 and 79.0, respectively. 

\begin{figure}[!t] 
    \centering
    \includegraphics[trim=0 {0.02\linewidth} 0 0, clip, width=1.0\linewidth]{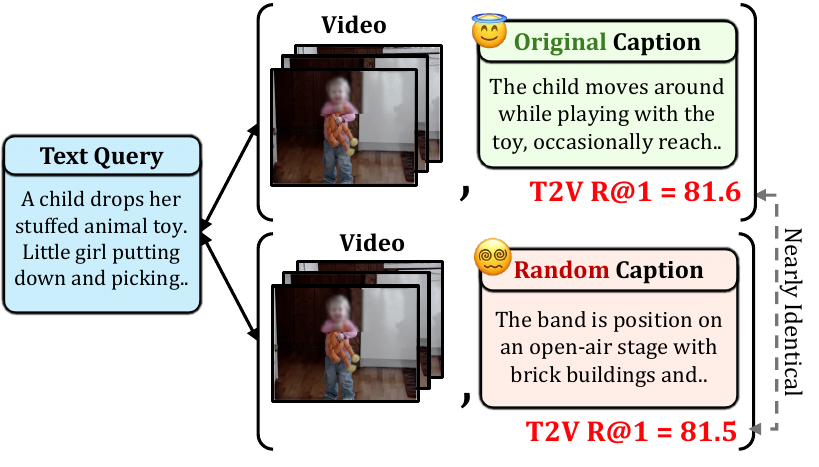}
    \caption{T2V retrieval with the \textit{original} descriptive caption (video-to-caption retrieval R@1 of 90.7) compared to the \textit{random} one. 
    Nearly identical performance suggests that the model fails to effectively leverage the auxiliary knowledge.}
    \label{supp-fig:observation}
\end{figure}

\begin{table*}[!h]
    \centering
    \begin{adjustbox}{width=0.80\linewidth}
    \small
    \begin{tabular}{c|cc|cc|cc|cc}
        \toprule
         &\multicolumn{2}{c|}{DiDeMo}
         &\multicolumn{2}{c}{ActivityNet} 
         &\multicolumn{2}{c}{MSRVTT} 
         &\multicolumn{2}{c}{MSVD}\\
         
         & $\mathcal{M}_{\text{ret}}$ 
         & $\mathcal{M}_{\text{cap}}$ 
         & $\mathcal{M}_{\text{ret}}$ 
         & $\mathcal{M}_{\text{cap}}$ 
         &  $\mathcal{M}_{\text{ret}}$ 
         & $\mathcal{M}_{\text{cap}}$ 
         &  $\mathcal{M}_{\text{ret}}$ 
         & $\mathcal{M}_{\text{cap}}$ \\

        \midrule
        \midrule
        Learning rate 
            & 8e-5 & 8e-6 
            & 2e-5 & 8e-6 
            & 1e-4 & 8e-6 
            & 2e-4 & 8e-6 \\
        Warmup Epochs & 1 & 0.1 & 1 & 0.1 & 1 & 0.1 & 1 & 0.1 \\
        Epoch & 5 & 1 & 5 & 1 & 3 & 1 & 5 & 1 \\
        Batch Size & 32 & 8 & 32 & 8 & 512 & 8 & 32 & 8 \\
        LoRA $r$ & 8 & 64 & 8 & 64 & 8 & 64 & 8 & 64 \\
        LoRA $\alpha$ & 32 & 128 & 32 & 128 & 32 & 128 & 32 & 128 \\
        $\lambda_{i \neq j}$ & - & 0.3 & - & 0.1 & - & 0.3 & - & 0.1 \\
        $\beta$ & - & 0.1 & - & 0.1 & - & 0.1 & - & 0.1 \\
        $\gamma$ & - & 0.7 & - &  2.0 & - & 0.5 & - & 0.01 \\
        \bottomrule
    \end{tabular}
    \end{adjustbox}
    \caption{Training hyperparameters for $\mathcal{M}_{\text{ret}}$ with $\mathcal{L}_{\text{ret}}$ and $\mathcal{M}_{\text{cap}}$ with $\mathcal{L}_{\text{DG-DPO}}$.
    }
    \label{supp:hparams_tab}
    \vspace{-0.2cm}
\end{table*}

\section{Implementation Details}
\label{supp:Implementation}
\paragraph{Training Details for retrieval.}
For training an MLLM-based model for retrieval, we adopt the recent MLLM of VideoChat-Flash-7B~\cite{li2024videochat}.
The baseline model is equipped with a visual encoder of UMT-L~\cite{li2023unmasked} and an LLM of Qwen2~\cite{yang2024qwen2}.
For each benchmark, we only train the linear projection layer while adopting LoRA~\cite{hu2021lora} for fine-tuning the model for efficiency.
We adopt 16 frames per video for all datasets.
All the experiments were done using 8 NVIDIA H100 80GB GPUS.

\paragraph{Prompts for text-video retrieval.}
We built several different models capable of implementing text-video retrieval.
For the model trained with the loss of $\mathcal{L} =-\log P(y|\mathbf{v}, \mathbf{t})$, which is the baseline text-video retrieval model that does not accept auxiliary caption as input, we adopted the prompt of ``\textit{Caption: \texttt{[text query]}. Does the above video match the caption? True or False}''.
Note that we utilized the word `Caption' for referencing the text query that is different from the auxiliary caption that we dealt with in this paper.
For training the model with the loss of $\mathcal{L}= -\log P(y|\mathbf{v}, \mathbf{c}, \mathbf{t})$, which is capable of adopting the auxiliary caption for training, we use the prompt of ``\textit{Video description: \texttt{[caption]}. Caption: \texttt{[text query]}. Based on the video and its description, is the video relevant to the caption? Answer True or False}.''
Again, to clarify, the `video description' corresponds to the auxiliary caption dealt with in the paper, whereas the `caption' refers to the text query (retrieval target).

\paragraph{Training details for captioning.}
\label{sec-supp:training-detail}
We adopt LLaVA-OneVision-7B~\cite{li2024llava} for training the captioning model with Direct Preference Optimization.
Similar to MLLM-based retrieval model finetuning, we adopt LoRA~\cite{hu2021lora} for parameter-efficient finetuning. 
LLaVA-Onevision consists of Qwen2~\cite{yang2024qwen2} as the LLM, and SigLIP vision encoder~\cite{zhai2023sigmoid}. 
We adopt 16 frames per video for all datasets.
All the experiments were done using 8 NVIDIA H100 80GB GPUS. 
Note that $\lambda_{i,j}$ in Eq.~\ref{eq:DG-DPO} denotes the weighting factor that balances the contribution between local pairs $(i=j)$ and cross-group pairs $(i \neq j)$.
Specifically, we denote the former as $\lambda_{i = j}$ and the latter as $\lambda_{i \neq  j}$.
The value of $\lambda_{i \neq  j}$ adopted for our experiments are in Tab~\ref{supp:hparams_tab}, and $\lambda_{i = j}$ is $(1-\lambda_{i \neq  j})$.

\paragraph{Prompt for captioning.}
We empirically explored several ways of generating the caption for the dataset for zero-shot.
Simply prompting the captioning model to generate a \textit{detailed} description about the video will cause the model to generate a very long paragraph for the given video.
Hence, we utilized the prompt of ``\textit{Describe this video in detail with three sentences.}''.

\paragraph{Inference details for retrieval.}
MLLM-based retrieval models are adopted as a re-ranker~\cite{lin2024mm, liu2024lamra, miech2021thinking}, benefiting from the ability to jointly attend to both visual and textual data.
Hence, based on the InternVideo-1B~\cite{wang2024internvideo2} similarity computed between the video and the text query, we retrieve the top-16 candidates for re-ranking. 
Finally, we weight the output scores of the two models following the protocol of ~\citet{miech2021thinking}.

\paragraph{Inference details for captioning.}
To construct the dataset $\mathcal{D}_{\text{DG-DPO}}$, we sample $k=3$ captions per video using a generation temperature of 0.2, following the settings provided by LLaVA-OneVision~\cite{li2024llava}.
For the caption generation of evaluating the retrieval model, we sample $k=2$ captions per video.
For our experiments, we average the retrieval scores across both in order to account for the variability in caption generation and provide a more robust performance estimate.

\paragraph{Training strategy for $\mathcal{L}_{\text{DG-DPO}}$.}
To construct preference pairs from ranked retrieval results, we explore two strategies. 
First, we compute a global rank for each sample in the dataset and refer to these ranks within each batch to determine preference between video-caption pairs.
The first strategy treats the top half of the ranked samples (i.e., higher-ranked pairs) as \textit{chosen} and the bottom half as \textit{rejected}.
In contrast, the second strategy forms preference pairs by grouping the ranked indices into adjacent pairs, where the higher-ranked sample in each pair is treated as the \textit{chosen} one and the lower-ranked as the \textit{rejected}.
Empirically, we observe that the latter strategy yields greater performance improvements. 
We hypothesize that this is because it produces training pairs with relatively smaller marginal differences compared to the former approach, allowing the model to learn more nuanced preference signals.

\section{Hyperparameters}
In Tab.~\ref{supp:hparams_tab}, we report the hyperparameters adopted for training the retrieval model $\mathcal{M}_{\text{ret}}$, and the captioning model $\mathcal{M}_{\text{cap}}$, across the text-video retrieval dataset.

\section{Further Ablation on Role-Embeddings.}
\label{sec:supp-role-embedding}
\paragraph{Ablation on each component.}
We conduct further analysis on the role-embeddings for text-video retrieval on DiDeMo, evaluating with and without each function role embedding in Tab.~\ref{supp:ablation-role-embeddings}.
The results suggest that the model trained with $R_{\text{t}}$ results in higher V2T retrieval at R@1 (79.2 to 79.8), whereas the model trained with $R_{\text{c}}$ results in higher T2V retrieval at R@1 (81.6 to 82.6).
Notably, combining both role embeddings results in the best overall performance, achieving 83.1 R@1 in T2V and 79.6 R@1 in V2T. 
These findings highlight the importance of role-embeddings integrated for both heterogeneous textual inputs, enhancing the model’s ability to distinguish the auxiliary caption and the retrieval target, enabling the model to utilize more of the auxiliary caption for retrieval.

\paragraph{Ablation with DG-DPO optimized caption.}
In Tab.~\ref{supp-tab:ablation-role-embeddings-finetuned}, we analyze the effectiveness of retrieval role-embeddings when adopting the caption that is generated by our DG-DPO optimized captioning model, similar to Tab.~\ref{tab:ablation-role-embeddings} in the main.
As shown, when replacing the caption with a random one, the model shows a minimal performance drop of 0.7 in R@1 on average (compare rows 1 and 2), whereas the model trained with role-embeddings shows higher sensitivity to the quality of the caption with a notable 1.6 point change in performance of R@1 (compare rows 3 and 4). 
Moreover, while adopting the same caption, the model trained with the role-embeddings yields superior performance, for instance, 85.1 compared to the baseline of 84.9 (compare rows 2 and 4).
\begin{table}[!t]
    \centering
    \renewcommand{\arraystretch}{1.2}
    \begin{adjustbox}{width=1.0\linewidth}
    \begin{tabular}{l|c|c|ccc|ccc}
        \toprule
        \textbf{Train} & &  & \multicolumn{3}{c|}{\textbf{Text-to-Video}} &\multicolumn{3}{c}{\textbf{Video-to-Text}} \\
        $\mathcal{L}_{\text{ret}}(\cdot)$ & $\mathbf{R}_{\text{c}}$ & $\mathbf{R}_{\text{t}}$  & R@1 & R@5 & R@10  & R@1 & R@5 & R@10 \\
        \hline
        \hline
        $(\mathbf{v}, \mathbf{c}, \mathbf{t})$
            & \ding{55} & \ding{55} & 81.6 & 94.3 & 95.9 & 79.2 & 94.7 & 96.7  \\
        \hline
        $(\mathbf{v}, \mathbf{c}, \mathbf{t}+\mathbf{R}_{\text{t}})$
            & \ding{55} & \ding{52} & 81.2 & 94.5 & 95.6 & 79.8 & 94.3 & 96.5  \\
        \hline
        $(\mathbf{v}, \mathbf{c}+\mathbf{R}_{\text{c}}, \mathbf{t})$
            & \ding{52} & \ding{55} & 82.6 & 94.4 & 95.9 & 79.6 & 94.6 & 96.6  \\
        \hline
        $(\mathbf{v}, \mathbf{c}+\mathbf{R}_{\text{c}}, \mathbf{t}+\mathbf{R}_{\text{t}})$
            &  \cellcolor{lightblue} \ding{52} &  \cellcolor{lightblue} \ding{52} &  \cellcolor{lightblue} 83.1 &  \cellcolor{lightblue} 94.4 & 
            \cellcolor{lightblue} 96.2 & 
             \cellcolor{lightblue} 79.6  & 
              \cellcolor{lightblue} 94.6 & 
               \cellcolor{lightblue} 96.6 \\
        \bottomrule
    \end{tabular}
    \end{adjustbox}
    \caption{\textbf{Ablation on each component of the role-embedding.}
    Note that we adopt the zero-shot captions with the standard inference strategy.
    }
    \label{supp:ablation-role-embeddings}
    \vspace{-0.3cm}
\end{table}

\begin{table}[!t]
    \centering
    \renewcommand{\arraystretch}{1.2}
    \setlength{\tabcolsep}{2pt}
    \begin{adjustbox}{width=1.0\linewidth}
    \begin{tabular}{c|c|c|ccc|ccc|c}
        \toprule
        \textbf{Train Input} & \multirow{2}{*}{\textbf{R$_{\text{c}}$}, \textbf{R$_{\text{t}}$}} & \textbf{Inf.}  & \multicolumn{3}{c|}{\textbf{Text-to-Video}} & \multicolumn{3}{c|}{\textbf{Video-to-Text}} & \multirow{2}{*}{\shortstack{$\text{Avg.}$ \\ $\Delta$}}\\
        $\mathcal{L}_{\text{ret}}(\cdot)$ & & cap. $\mathbf{c}$  & R@1 & R@5 & R@10  & R@1 & R@5 & R@10 & \\
        \midrule
        \midrule
        \multirow{2}{*}{$(\mathbf{v}, \mathbf{c}, \mathbf{t})$} 
            &  \ding{55} & rand. 
            & 81.6 & 94.9 & 96.4 & 83.4 & 94.1 & 95.9 & - \\
            
            & \cellcolor{lightblue}  \ding{55} & \cellcolor{lightblue} ours & \cellcolor{lightblue} 84.9 & \cellcolor{lightblue} 95.4 & \cellcolor{lightblue} 96.2 & \cellcolor{lightblue} 82.1 & \cellcolor{lightblue} 95.3 & \cellcolor{lightblue} 96.4 & \cellcolor{lightblue} (+0.7) \\
        \midrule
        \multirow{2}{*}{$(\mathbf{v}', \mathbf{c}', \mathbf{t}')$} 
            & \ding{52} & rand. & 80.9 & 94.6 & 96.3 & 80.9 & 95.1 & 96.1 & - \\
            & \cellcolor{lightblue} \ding{52} & \cellcolor{lightblue} ours & \cellcolor{lightblue} 85.1 & \cellcolor{lightblue} 95.0 & \cellcolor{lightblue} 96.2 & 
            \cellcolor{lightblue} 82.5 & \cellcolor{lightblue} 95.2 & \cellcolor{lightblue} 96.3 & \cellcolor{lightblue} (+1.6) \\
        \bottomrule
    \end{tabular}
    \end{adjustbox}
    \caption{\textbf{Ablation on the Role-embeddings of $\mathcal{M}_{\text{ret}}$ with DG-DPO optimized caption.}
    We adopt our DG-DPO optimized captions, denoted with `our', with the contrastive inference strategy.
    `rand.' denotes \textit{random} caption and `Inf.' denotes the inference stage.
    $\mathbf{c}'$ and $\mathbf{t}'$ denote $\mathbf{c}+\mathbf{R}_{\text{c}}$ and $\mathbf{t}+\mathbf{R}_{\text{t}}$, respectively. 
    `Avg. $\Delta$' denotes an average change in R@k performance, between `rand' and `ours' captions.
    }
    \label{supp-tab:ablation-role-embeddings-finetuned}
\end{table}

\begin{table}[!t]
    \centering
    \begin{adjustbox}{width=1.0\linewidth}
    \renewcommand{\arraystretch}{1.0}
    \begin{tabular}{c|c c c|c c c}
        \toprule
          \multirow{3}{*}{$s(\mathbf{v}, \mathbf{c}, \mathbf{t})$} & \multicolumn{3}{c|}{\textbf{Text-to-Video}} & \multicolumn{3}{c}{\textbf{Video-to-Text}} \\
          & R@1 & R@5 & R@10  & R@1 & R@5 & R@10 \\
        \midrule
        \midrule
         $\log P(y^+|\mathbf{v}, \mathbf{c}, \mathbf{t})$ & 82.6 & 94.7 & 96.2 & 79.7 & 94.7 & 96.1  \\
         $\log P(y^-|\mathbf{v}, \mathbf{c}, \mathbf{t})$ & 84.9 & 95.0 & 96.2 & 82.3  & 95.1 & 96.2  \\
        \rowcolor{lightblue}
         $\log \displaystyle\frac{P(y^+|\mathbf{v}, \mathbf{c}, \mathbf{t})}{P(y^-|\mathbf{v}, \mathbf{c}, \mathbf{t})}$  & 85.1 & 95.0 & 96.2 & 82.5 & 95.2 & 96.3 \\
        \bottomrule
    \end{tabular}
    \end{adjustbox}
    \caption{\textbf{Comparison on the inference strategy.} 
    Retrieval performance on DiDeMo, where ${s}(\mathbf{v}, \mathbf{c}, \mathbf{t})$ denotes the relevance score adopted for the inference.}
    \label{supp:inference}
\end{table}
\paragraph{Analysis on the inference strategy.}
Tab.~\ref{supp:inference} explores the different inference strategies in MLLM retrieval, and we determine that our contrastive inference strategy yields the best result. 
The standard approach (row 1) results in significant performance degradation compared to those that adopt the probability of generating `\texttt{False}' (row 2 and 3).
Specifically, simply adopting $\log P(y^{-}|\mathbf{v}, \mathbf{c}, \mathbf{t})$ (row 2), results in +2.5\% increase in R@1 on average, and adopting $\log P(y^{+}|\mathbf{v}, \mathbf{c}, \mathbf{t}) - \log P(y^{-}|\mathbf{v}, \mathbf{c}, \mathbf{t})$ (row 3), results in +2.7\% increase.

\begin{table}[!t]
    \centering
    \begin{adjustbox}{width=0.97\linewidth}
    \begin{tabular}{l|c c c|c c c}
        \toprule
         & \multicolumn{3}{c|}{\textbf{Text-to-Video}} & \multicolumn{3}{c}{\textbf{Video-to-Text}} \\
        & R@1 & R@5 & R@10  & R@1 & R@5 & R@10 \\
        \midrule
        \multicolumn{7}{l}{\textit{Captioning Metric}} \\
        \hline
        \midrule
        
         \text{BLEU} & 84.1 & 95.0 & 96.3 & 82.3 & 94.7 & 96.4  \\
         \text{METEOR} & 83.8 & 94.9 & 96.6 & 82.8 & 94.9 & 96.3 \\
        \midrule
        \multicolumn{7}{l}{\textit{Retrieval Score ($s_p$)}} \\
        \hline
        \midrule
        \rowcolor{lightblue}
         $\log \displaystyle\frac{P(y^+|\mathbf{v}, \mathbf{c}, \mathbf{t})}{P(y^-|\mathbf{v}, \mathbf{c}, \mathbf{t})}$ & 85.0 & 95.0 & 96.4 & 82.4 & 95.0 & 96.4 \\
        \rowcolor{lightblue}
         $\log \displaystyle\frac{P(y^+|\mathbf{v}^{*}, \mathbf{c}, \mathbf{t})}{P(y^-|\mathbf{v}^{*}, \mathbf{c}, \mathbf{t})}$ & 85.1 & 95.0 & 96.2 & 82.5 & 95.2 & 96.3 \\
        \bottomrule
    \end{tabular}
    \end{adjustbox}
    \caption{\textbf{Comparison on adopting different preference scores $s_p$ for constructing $\mathcal{D}_{\text{DG-DPO}}$.}
    We report the retrieval performance on DiDeMo.
    Also, $\mathbf{v}^{*}$ denotes masked attention for video tokens.
    }
    \label{tab:relevance-score}
\end{table}
\begin{table}[t]
\centering
\begin{adjustbox}{width=0.8\linewidth}
\begin{tabular}{l|c|c}
\toprule
\textbf{Method} & \textbf{T2V R@1} & \textbf{V2T R@1} \\
\midrule
Baseline & 83.1 & 79.6 \\
$+ \mathcal{L}_{\text{SFT}}$ & 82.6 & 82.0 \\
$+ \mathcal{L}_{\text{DG-DPO}}$ (Batch=4) & 84.8 & 82.6 \\
$+ \mathcal{L}_{\text{DG-DPO}}$ (Batch=8) & 84.2 & \textbf{82.8} \\
$+ \mathcal{L}_{\text{DG-DPO}}$ (Batch=16) & 84.2 & 82.5 \\
$+ \mathcal{L}_{\text{DG-DPO}}$ (Batch=32) & \textbf{85.1} & 82.5 \\
\bottomrule
\end{tabular}
\end{adjustbox}
\caption{\textbf{Effect of DG-DPO with varying batch sizes.} We report R@1 for text-to-video (T2V) and video-to-text (V2T) retrieval.}
\label{tab:dg-dpo-batch}
\end{table}

\section{Further Ablation on DG-DPO}
\paragraph{Ablation on preference scores.}
\label{supp:preference_score}
In Tab~\ref{tab:relevance-score}, we compare retrieval performance while adopting different types of preference scores $s_p$ for constructing $\mathcal{D}_{\text{DG-DPO}}$.
We observe that directly using retrieval-based scores (rows 3 and 4) consistently outperforms traditional captioning metrics such as BLEU~\cite{papineni2002bleu} and METEOR~\cite{banerjee2005meteor}.
Specifically, adopting BLEU and METEOR leads to a performance drop of 1.0 and 1.3 points in T2V R@1, respectively.
For V2T, the impact of caption quality appears marginal, as captions primarily augment the video rather than the text query, contributing more significantly to improvements in T2V.

\paragraph{Ablation on DG-DPO batch size.}
\label{supp:batch_size}
To analyze the effect of batch size in preference optimization, we conduct an ablation study by applying DG-DPO with varying batch sizes during training. 
As shown in Table~\ref{tab:dg-dpo-batch}, all DG-DPO variants outperform the baseline and SFT-only models, demonstrating the effectiveness of direct preference optimization. 
Notably, performance improves consistently as the batch size increases, with the best results observed at a batch size of 32, achieving 85.1 R@1 in text-to-video and 82.5 R@1 in video-to-text retrieval. 
This highlights the benefit of utilizing diverse preference pairs per update and suggests that larger batch sizes help the model better capture fine-grained semantic preferences aligned with retrieval objectives.

\section{Out-of-Domain Generalization}
\begin{table}[!t]
    \centering
    \begin{adjustbox}{width=1.0\linewidth}
    \begin{tabular}{c|c|c|ccc|ccc}
        \toprule
        \multirow{2}{*}{\shortstack{$\textbf{Training}$ \\ \textbf{Dataset}}} &
        \multirow{2}{*}{\shortstack{$\textbf{Test}$ \\ \textbf{Dataset}}} &
        \multirow{2}{*}{\shortstack{$\textbf{Preference}$ \\ \textbf{score}}} &
        \multicolumn{3}{c|}{\textbf{T2V}} & \multicolumn{3}{c}{\textbf{V2T}} \\
        & & & R@1 & R@5 & R@10 & R@1 & R@5 & R@10 \\
        \midrule
        DiDeMo & MSRVTT & BLEU       & 63.9 & 82.9 & 87.2 & 63.7 & 83.0 & 87.2 \\
        DiDeMo & MSRVTT & retrieval  & 64.2 & 83.8 & 88.5 & 63.9 & 83.1 & 87.3 \\
        MSRVTT & MSRVTT & retrieval  & 64.1 & 83.8 & 88.8 & 63.8 & 83.0 & 87.3 \\
        \bottomrule
    \end{tabular}
    \end{adjustbox}
    \caption{\textbf{Cross-dataset generalization of DG-DPO captions.}
    To evaluate out-of-domain generalization of $\mathcal{M}_{\text{cap}}$ with DG-DPO on the DiDeMo dataset and test retrieval performance on MSRVTT, which differs in video length and query granularity.}
    \label{tab:ood}
\end{table}

To assess the out-of-domain generalization ability of our method, we conducted a cross-dataset evaluation by training the captioner with DG-DPO on DiDeMo and evaluating retrieval performance on the MSRVTT benchmark, of which the former contains longer videos with more moment-specific queries, whereas the latter contains shorter videos with queries that refer to the entire video.
As shown in Tab~\ref{tab:ood}, DG-DPO-trained captions generalize well to unseen datasets, even outperforming the performance when trained with the specific dataset (see row 2 and row 3).
In addition, retrieval score-based optimized captions outperform BLEU-optimized captions even when trained out-of-domain. 
This confirms that DG-DPO captures meaningful retrieval-aligned semantics rather than overfitting to domain-specific biases.

\section{Further Qualitative Results}
\label{supp:qualitatives}
\begin{figure*}[!h] 
    \centering
    \begin{subfigure}[h]{\linewidth}
        \includegraphics[width=\linewidth]{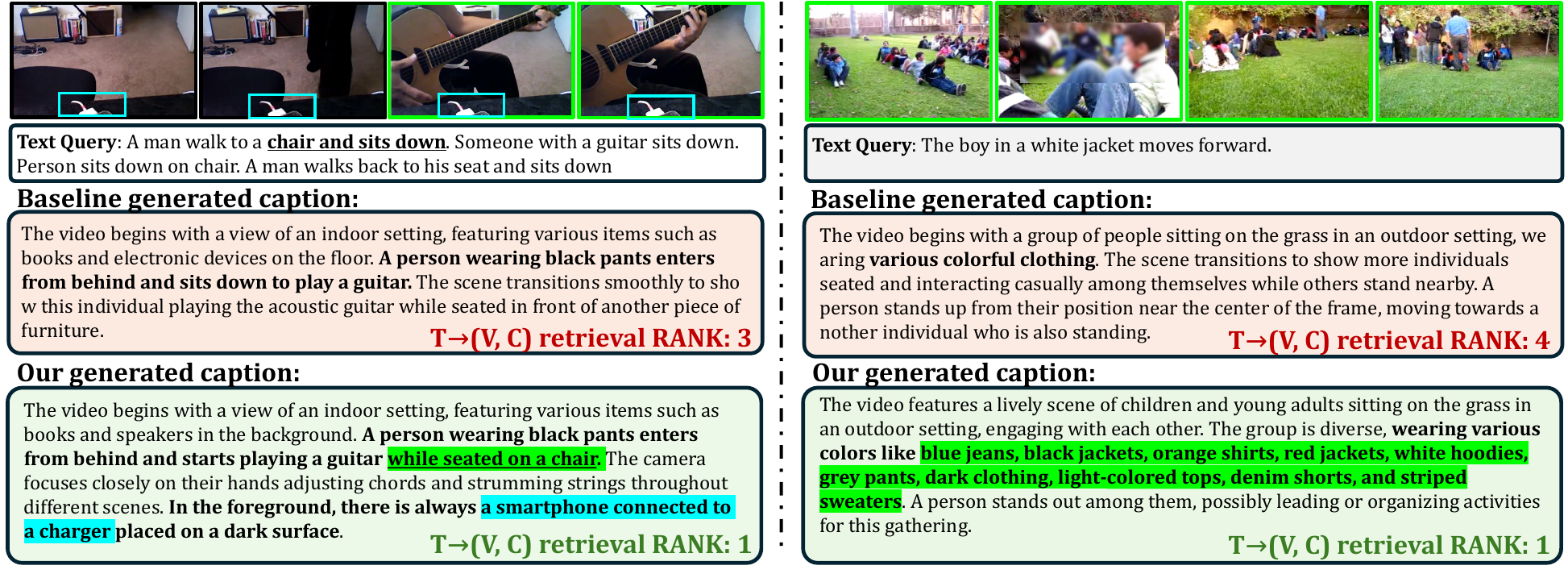}
        \caption{Qualitative example on the effect of generated video caption in \textbf{text-to-video} retrieval on DiDeMo.}
        \label{fig:supp-didemo-qual1}
    \end{subfigure}
    \begin{subfigure}[h]{\linewidth}
        \includegraphics[width=\linewidth]{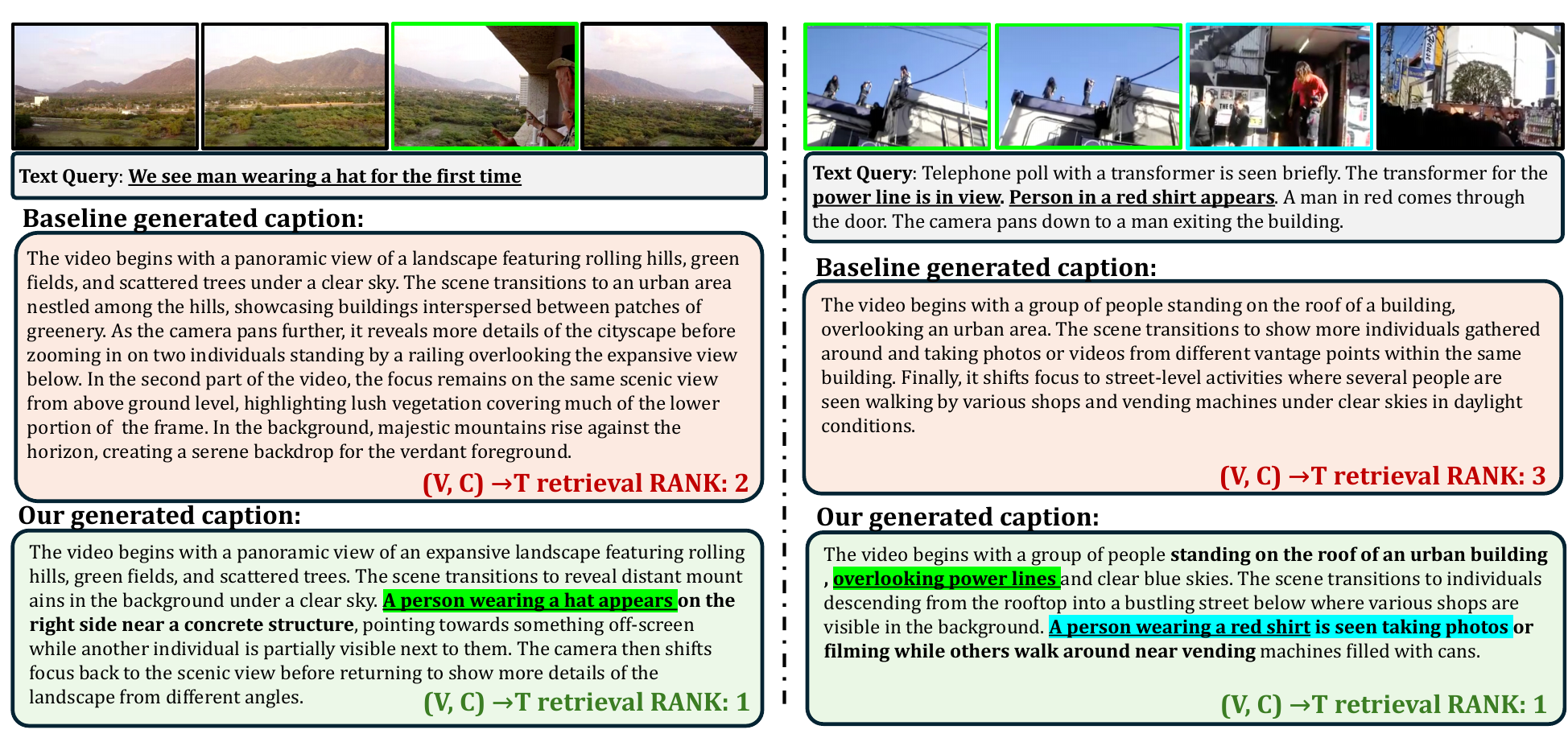}
        \caption{Qualitative example on the effect of generated video caption in \textbf{video-to-text} retrieval on DiDeMo.}
        \label{fig:supp-didemo-qual2}
    \end{subfigure}
    \begin{subfigure}[h]{\linewidth}
        \includegraphics[width=\linewidth]{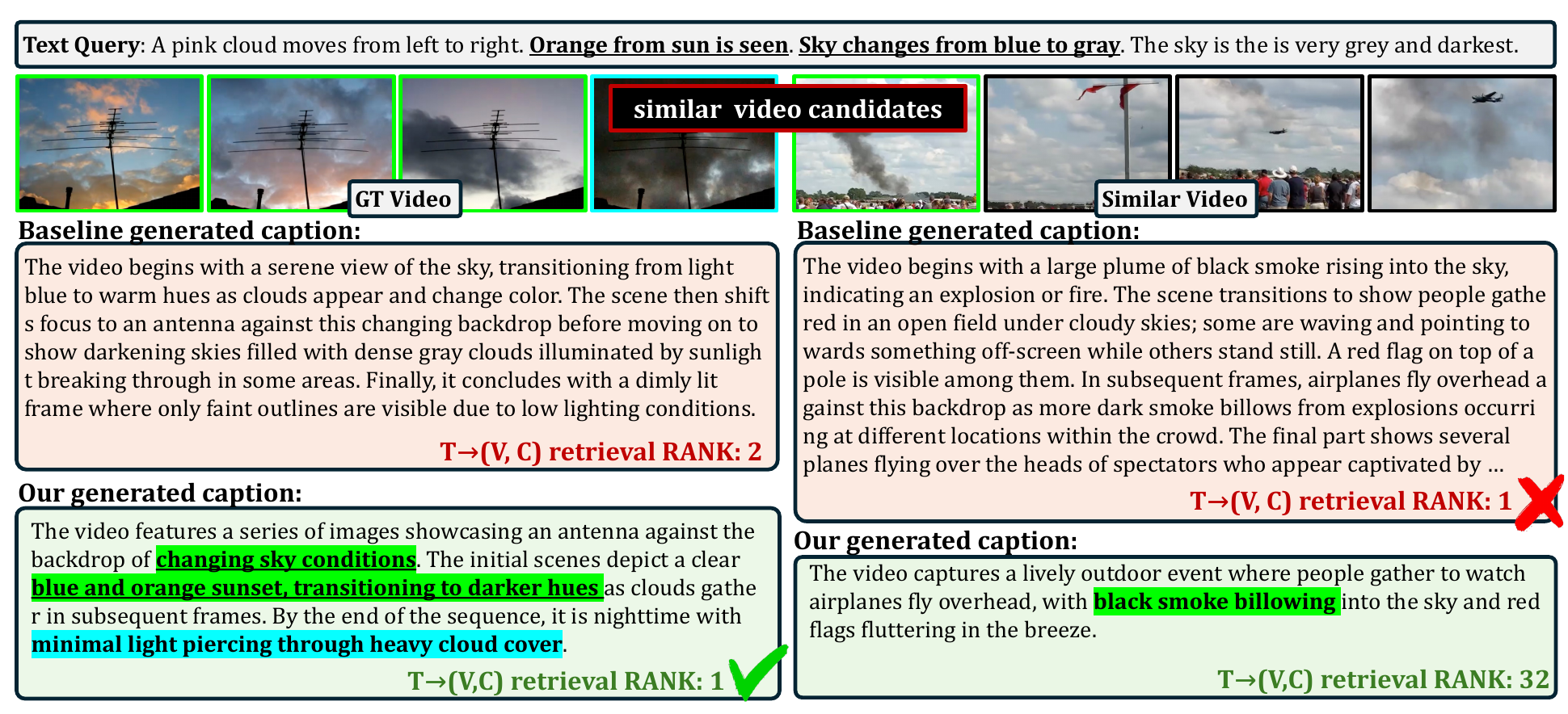}
        \caption{Qualitative example on the effect of generated video caption in \textbf{text-to-video} retrieval on DiDeMo for Challenging Retrieval Cases.}
        \label{fig:supp-didemo-qual3}
    \end{subfigure}
    \caption{\textbf{Further qualitative example of video captioning.}
    Comparison of the predictions of the caption generated by the zero-shot captioning model (Baseline) with our model trained with DG-DPO (Ours) on DiDeMo.
    The highlighted parts depict the fine-grained detail generated by our model, which is not provided in the caption generated by the baseline.
    The border color of each video frame corresponds to the caption highlighted, and the underline denotes the details that closely match the given query.}
    \label{fig:supp-qual-didemo}
\end{figure*}

\begin{figure*}[!h] 
    \centering
    \begin{subfigure}[h]{\linewidth}
        \includegraphics[width=\linewidth]{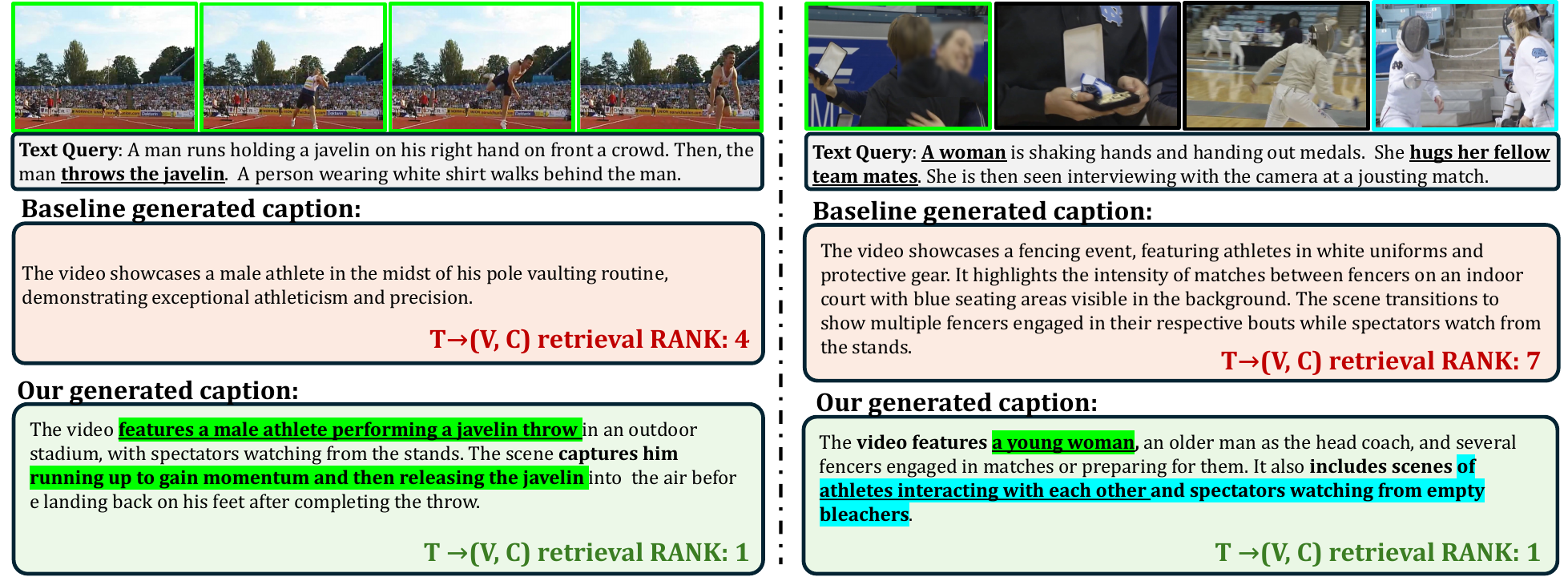}
        \caption{Qualitative example on the effect of generated video caption in \textbf{text-to-video} retrieval on ActivityNet.}
        \label{fig:supp-act-qual1}
    \end{subfigure}
    \begin{subfigure}[h]{\linewidth}
        \includegraphics[width=\linewidth]{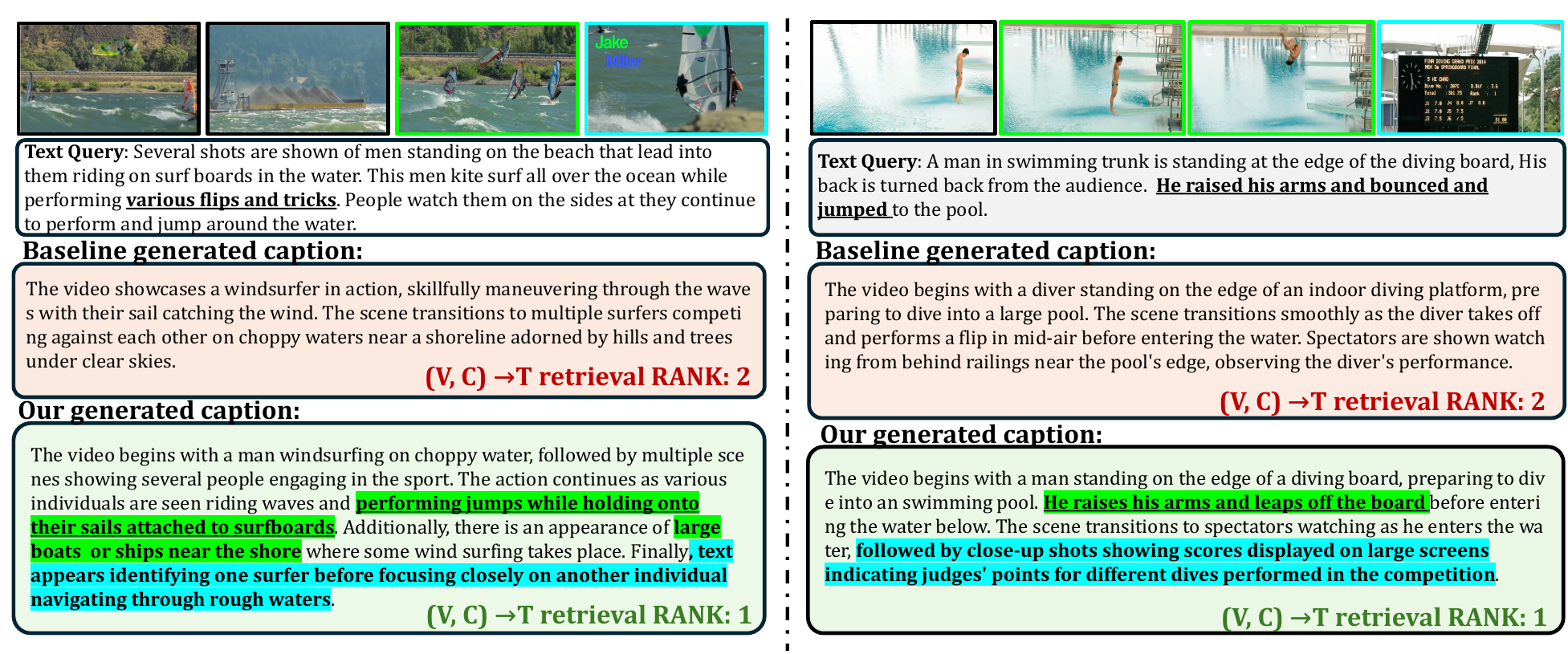}
        \caption{Qualitative example on the effect of generated video caption in \textbf{video-to-text} retrieval on ActivityNet.}
        \label{fig:supp-act-qual2}
    \end{subfigure}
    \begin{subfigure}[h]{\linewidth}
        \includegraphics[width=\linewidth]{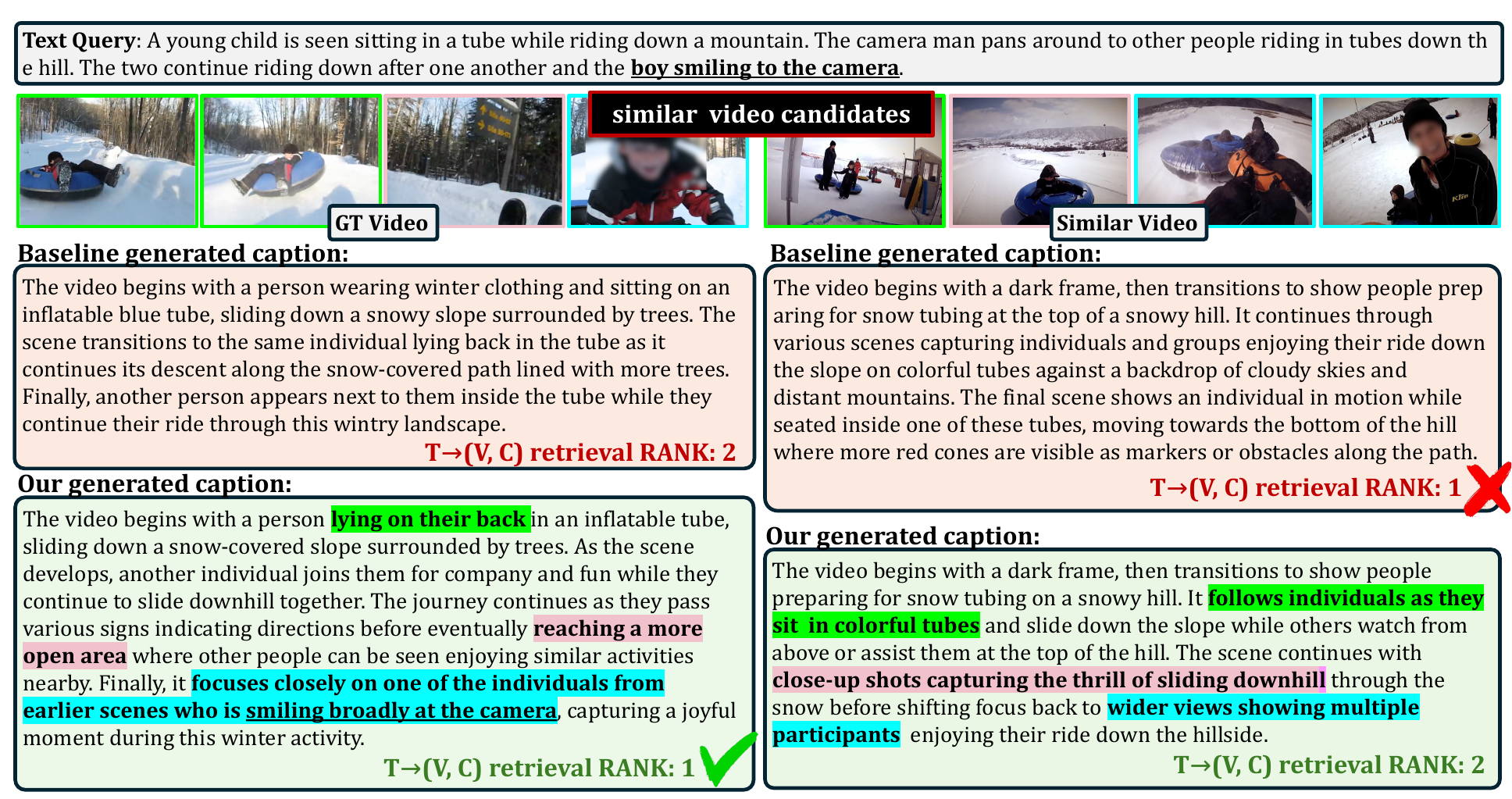}
        \caption{Qualitative example on the effect of generated video caption in \textbf{text-to-video} retrieval on ActivityNet for Challenging Retrieval Cases.}
        \label{fig:supp-act-qual3}
    \end{subfigure}
    \caption{\textbf{Further qualitative example of video captioning.}
    Comparison of the predictions of the caption generated by the zero-shot captioning model (Baseline) with our model trained with DG-DPO (Ours) on ActivityNet.
    The highlighted parts depict the fine-grained detail generated by our model, which is not provided in the caption generated by the baseline.
    The border color of each video frame corresponds to the caption highlighted, and the underline denotes the details that closely match the given query.}
    \label{fig:supp-qual-act}
\end{figure*}

In Fig~\ref{fig:supp-qual-didemo}, Fig~\ref{fig:supp-qual-act}, we further show the qualitative results between captions from the baseline and our model on DiDeMo and ActivityNet, focusing on those benchmarks that require fine-grained detail for retrieval.
All figures (a) depict examples for text-to-video retrieval, whereas (b) depict examples of video-to-text retrieval. 
(c) illustrates cases where, when adopting a baseline (zero-shot) caption, the model confuses between the two similar videos for text-to-video retrieval, whereas with DG-DPO optimized captions, the fine-grained details successfully lead to discriminating between the videos. 
Overall, our model trained with the DG-DPO of our CaRe-DPO enables the retrieval model to better align with the text-video retrieval task.

\end{document}